\documentclass{article}
\usepackage[utf8]{inputenc}
\usepackage[T1]{fontenc}
\usepackage{graphicx}
\usepackage{hyperref}
\usepackage{subcaption}
\usepackage{mathtools}
\usepackage{booktabs}
\usepackage{bm}
\usepackage{tcolorbox}
\usepackage{multicol,multirow}
\usepackage{natbib}
\usepackage{comment}
\usepackage{booktabs}

\newcommand{\ie}{\textit{i.e.}}

\usepackage{enumerate}   
\usepackage{amsmath}
\usepackage{multirow}
\usepackage{bbm}
\usepackage{tikz}

\definecolor{mygreen}{rgb}{0,0.6,0}
\usetikzlibrary{positioning,arrows,decorations,decorations.markings,shadows,positioning,arrows.meta,matrix,fit}

\usetikzlibrary{positioning,arrows}
\usepackage{wrapfig}

\usepackage{amsmath}
\usepackage{appendix}
\usepackage{amssymb}
\usepackage[cal=boondoxo]{mathalfa}
\newcommand{\cu}{\mathcal{U}}

\usepackage{amsthm}
\usepackage[capitalise]{cleveref}
\Crefname{assumption}{Assumption}{Assumptions}
\newcommand{\traj}{\mathcal{T}}

\theoremstyle{plain}
\newtheorem{theorem}{Theorem}
\newtheorem{lemma}{Lemma}

\theoremstyle{definition}

\newtheorem{definition}{Definition}
\newtheorem{remark}{Remark}

\newcommand{\bG}{\mathbb{G}}

\def\Cramer{Cram\'{e}r}
\newcommand{\Tcal}{\mathcal{T}}

\newcommand{\bN}{\mathcal{N}}
\newcommand{\TI}{\mathrm{TI}}
\newcommand{\MO}{\mathrm{MO}}

\newcommand{\cm}{\mathcal{M}}
\newcommand{\cN}{\mathcal{N}}

\newcommand{\bigO}{\mathcal{O}} %
\newcommand{\cj}{\mathcal{J}}
\newcommand{\cjj}{\mathcal{j}}

\renewcommand{\P}{\mathbb{P}}

\newcommand{\op}{\mathrm{o}_{p}}
\newcommand{\E}{\mathbb{E}}

\newcommand{\pa}{\mathrm{\pa}}

\newcommand{\var}{\mathrm{var}}

\newcommand{\rd}{\mathrm{d}}

\newcommand{\prns}[1]{\left(#1\right)}
\newcommand{\braces}[1]{\left\{#1\right\}}
\newcommand{\bracks}[1]{\left[#1\right]}

\newcommand{\abs}[1]{\left|#1\right|}

\newcommand{\epol}{\pi^\mathrm{e}}
\newcommand{\bpol}{\pi^\mathrm{b}}

\newcommand{\Scal}{\mathcal{S}}
\newcommand{\Acal}{\mathcal{A}}

\newcommand{\Lcal}{\mathcal{L}}

\renewcommand{\eqref}[1]{(\ref{#1})}

\newcommand{\RN}[1]{%
  \textup{\uppercase\expandafter{\romannumeral#1}}%
}

\usepackage{color,graphicx,lscape,longtable}

\def\boxit#1{\vbox{\hrule\hbox{\vrule\kern6pt\vbox{\kern6pt#1\kern6pt}\kern6pt\vrule}\hrule}}

\usepackage{xcolor}
\newcount\Comments  
\newcommand{\kibitz}[2]{\ifnum\Comments=1\textcolor{#1}{#2}\fi}

\newcommand{\nk}[1]{\kibitz{blue}{[NK: #1]}}

\usepackage{microtype}
\usepackage{graphicx}
\usepackage{algorithm}
\usepackage{algorithmic}

\usepackage{url} 
\allowdisplaybreaks

\title{Efficient Evaluation of Natural Stochastic Policies in Offline Reinforcement Learning}
\author{Nathan Kallus \\ Department of Operations Research and Information Engineering\\  and Cornell Tech
       Cornell University 
       \and 
      \and
       Masatoshi Uehara \footnote{Corresponding author \href{mailto:mu223@cornell.edu}{mu223@cornell.edu}} \\ Department of Computer Science and Cornell Tech\\
       Cornell University 
       }

\date{}

\begin{document}

\maketitle 

\begin{abstract}
We study the efficient off-policy evaluation of natural stochastic policies, which are defined in terms of deviations from the behavior policy. This is a departure from the literature on off-policy evaluation where most work consider the evaluation of explicitly specified policies. Crucially, offline reinforcement learning with natural stochastic policies can help alleviate issues of weak overlap, lead to policies that build upon current practice, and improve policies' implementability in practice. Compared with the classic case of a pre-specified evaluation policy, when evaluating natural stochastic policies, the efficiency bound, which measures the best-achievable estimation error, is inflated since the evaluation policy itself is unknown. In this paper we derive the efficiency bounds of two major types of natural stochastic policies: tilting policies and modified treatment policies. We then propose efficient nonparametric estimators that attain the efficiency bounds under very lax conditions. These also enjoy a (partial) double robustness property.
\end{abstract}

\noindent%
\vfill


\section{Introduction} \label{sec:intro}

In many emerging application domains for reinforcement learning (RL), exploration is highly limited and simulation unreliable, such as in healthcare \citep{Kosorok2015,gottesman2019guidelines}. In these domains, we must use offline RL, where we evaluate and learn new sequential decision policies from existing observational data \citep{MurphyS.A.2003Odtr,Robins2004,Zhang2013,pmlr-v97-bibaut19a,Kallus2019IntrinsicallyES,ChowYinlam2019DBEo}. A key task in offline RL is that of off-policy evaluation (OPE), in which we evaluate a new policy from data logged by another behavior policy. In many applications such as mobile health, the horizon is often long and possibly infinite \citep{Audrey2018,LuckettDanielJ.2018EDTR,Liao2020,ShiC2020SIot}. In such settings, the na\"ive sequential importance sampling estimator suffers from the curse of horizon \citep{Liu2018} in the sense that its mean squared error (MSE) grows exponentially in the horizon. Recent work in OPE \citep{KallusUehara2019,KallusNathan2019EBtC} has shown how \emph{efficiently} leveraging problem structure, such as Markovianness and time-homogeneity, can significantly improve OPE and address issues such as the curse of horizon. 

In most of the literature on OPE, including the above, the policy to be evaluated is \emph{pre-specified}, that is, it is a given and known function from states to a distribution over actions. In a departure from this, in this paper we consider the evaluation of \emph{natural stochastic policies}, which may depend on the \emph{natural} value of the action, that is, the treatment that is observed in the data without intervention \citep{MunozIvanDiaz2012PICE,Shpitser2012,HaneuseS.2013Eote,YoungJessicaG2014Ieaa,Young2019,SWIGs,Diaz2018}. Specifically, we consider policies defined as \emph{deviations} from the behavior policy that generated the observed data.

There are two primary advantages to natural stochastic policies. A first advantage is \emph{implementability}. Subjects are often unable or reluctant to undertake an assigned treatment if the deviation from the treatment they would have naturally undertaken is large. For example, consider intervening on leisure-time physical activity to reduce mortality among the elderly \citep[as in][]{Diaz2018}. An evaluation policy assigning $a+\delta$ minutes of weekly activity to an individual whose current physical activity level is $a$ (i.e., the natural value) would be a realistic intervention for small to moderate $\delta$. On the other hand, evaluation policies assigning any arbitrary level of physical activity level ignoring the current level of physical activity is unrealistic and rarely implementable. Another example is intervening on air pollution levels to improve the health of children \citep[as in][]{DiazIvan2013Atce}. A possible evaluation policy is enforcing the pollution levels below a certain cutoff point if the observed pollution level (i.e., the natural value) exceeds the threshold. A second advantage is that we can relax or more easily satisfy the positivity assumption, which requires some overlap between the evaluation and behavior policies and is fundamentally necessary for OPE. Often, we cannot know a priori whether the positivity assumption is satisfied for a given intervention in an observational study or how good is the overlap. We can, however, easily consider policies that only deviate slightly from the behavior policy, ensuring a good overlap and reliable evaluation by design.

In this paper, we derive efficiency bounds and develop efficient estimators for two major types of natural stochastic policies: \emph{tilting policies} and \emph{modified treatment policies}. The efficiency bounds quantify 
the statistical limits to evaluation by showing what is the best-achievable MSE asymptotically. We study how much the efficiency bounds inflate in comparison with the case of a pre-specified evaluation policy (see \cref{tab:comparison}). Our central message is 
that the order of the efficiency bound is surprisingly the same as in the case of a pre-specified policy. Importantly, this implies the curse of the horizon is avoidable since the horizon dependence is polynomial, while good overlap can be achieved by design. We then develop efficient estimation methods achieving these efficiency bounds under 
lax conditions. We also demonstrate how methods that are efficient for pre-specified policies \citep{KallusUehara2019,KallusNathan2019EBtC} break. Unlike the pre-specified case, the estimator has an interesting \emph{partial} double robustness property, which is a different from the usual double robustness \citep{robins99,jiang2016,KallusUehara2019}, due to a special bias structure. 

{The article is organized as follows. \cref{sec:setup} sets up the problem and definitions and discusses related literature. \cref{sec:bounds} establishes the efficiency bounds for OPE of natural stochastic policies in time-varying Markov decision processes (TMDPs). \cref{sec:tmdp} develops efficient estimators. \cref{sec:nmdp} provides the efficiency bound and efficient estimators when we remove the Markovian assumption. \cref{sec:mdp} provides the efficiency bound and efficient estimators when we additionally impose time invariance on the decision process. \cref{sec:empirical} studies the performance of our approach empirically. \cref{sec:conc} provides concluding remarks on our findings.}

\begin{table}[t!]
    \centering
        \caption{Comparison of efficiency bounds for OPE under different types of policies in a TMDP.}
\label{tab:comparison}
  {\small
    \begin{tabular}{llll}\toprule
    & \multirow{2}{*}{Form of $\epol_t(a_t\mid s_t)$}  & Inflation of the bound  & Order of  \\ 
    &  & compared with (a)  &   the bound  \\ \midrule
   (a) Pre-specified  & Explicit function   &  0   & $CC'R^2_{\max}H^2$  \\
   (b) \textbf{Tilting} & $\frac{u_t(a_t)\bpol_t(a_t\mid s_t)}{\int u_t(\tilde a_t)\bpol_t(\tilde a_t\mid s_t)\mathrm{d}\tilde a_t}$     &   $\sum_{t=0}^H \E[\var[\mu_t q_t \mid s_t ]]$ & $CC'R^2_{\max}H^2$    \\
   (c) \textbf{Modified treatment} &   
   $\bpol_t(\tilde\tau_t(s_t,a_t)\mid s_t)\tilde\tau'_t(s_t,a_t)$
   &  $\sum_{t=0}^H \E[\mu^2_t\var[q^{\tau}_{t+1}\mid s_{t+1}]]$ &$CC'R^2_{\max}H^2$   \\\bottomrule
    \end{tabular}
    }
\end{table}

\section{Setup and Background}\label{sec:setup}

We setup the problem and notation and summarize the literature on OPE and natural stochastic policies.

\subsection{Problem Setup and Definitions} \label{sec:notationanddefs}

Consider an $H$-long time-varying Markov decision process (TMDP), with states $s_t\in\mathcal S_t$, actions $a_t\in\mathcal A_t$, rewards $r_t\in\mathbb R$, initial state distribution $p_1(s_1)$, transition distributions $p_{t+1}(s_{t+1}\mid s_t,a_t)$, and reward distributions $p_t(r_{t}\mid s_t,a_t)$, for $t=1,\dots,H$. 
A policy $(\pi_t(a_t\mid s_t))_{t\leq H}$ induces a distribution over trajectories $\traj=(s_1,a_1,r_1,\dots,s_T,a_H,r_H,s_{H+1})$:
\begin{align} \label{eq:trajdist_mdp}
p_\pi(\traj)= p_1(s_1)\prod_{t=1}^H\pi_t(a_t\mid s_t)p_t(r_t\mid s_t,a_t)p_{t+1}(s_{t+1}\mid s_t,a_t).
\end{align}
Given an evaluation policy $\epol$, which we consider as \textit{unknown} in this paper, we are interested in its value, $J=\E_{p_{\epol}}\bracks{\sum_{t=1}^Hr_t}$, where the expectation is taken with respect to (w.r.t.) the density induced by the evaluation policy, $p_{\epol}$. In the \emph{off-policy} setting, our data consists of trajectory observations from some fixed policy, $\pi^b$, known as the \emph{behavior policy}:
\begin{equation}\tag{Off-policy data}\label{eq:offpolicydata}
\traj^{(1)},\dots,\traj^{(n)}\sim p_{\pi^b},\quad\traj^{(i)}=(S^{(i)}_1,A^{(i)}_1,R^{(i)}_1,\cdots,S^{(i)}_H,A^{(i)}_H,R^{(i)}_H). 
\end{equation}
In observational studies, as we consider herein, $\bpol$ is \emph{unknown}, and the observed action $A^{(i)}_j$ is considered the natural value of the action in the sense that it is the one naturally observed in the absence of our intervention. Our goal is to estimate $J$ from the observed data $\{\traj^{(i)}\}_{i=1}^n$.\footnote{Although we do not explicitly use a counterfactual notation 
this is the same as the counterfactual value of following $\epol$ instead of $\bpol$ if we had used potential outcomes and assumed the usual sequential ignorability and consistency assumptions \citep{ErtefaieAshkan2014CDTR,LuckettDanielJ.2018EDTR}. }

We define the following variables. Let $q_t  =\E_{p_{\epol}}\bracks{\sum_{k=t}^Hr_t\mid s_t,a_t},\,v_t=\E_{p_{\epol}}\bracks{\sum_{k=t}^Hr_t\mid s_t}$ be the $q$- and $v$-functions for $\epol$. Further, let the instantaneous, cumulative, marginal state, and marginal state-action density ratios be, respectively, $
\eta_t =\frac{\pi_{t}^e(a_t\mid s_t)}{\pi_{t}^b(a_t\mid s_t)},\,\lambda_t=\prod_{k=1}^t \eta_k,\, 
w_t =\frac{p_{\pi^e}(s_t)}{p_{\pi^b}(s_t)},\mu_t=\eta_tw_t$, where $p_{\pi}(s_t)$ is a marginal density at $s_t$ under $p_\pi$. 
We assume throughout the paper that $0\leq r_t\leq R_{\max},\,\eta_{t} \leq C,\,w_t \leq C'$, $\forall t\leq H$.

Given trajectory data, $\traj^{(1)},\dots,\traj^{(n)}$, we define the \emph{empirical expectation} as
$\P_nf=\frac1n\sum_{i=1}^nf(\traj^{(i)})$. Unless otherwise noted, all expectations, variances, and probabilities are w.r.t. $p_{\bpol}$. Define the $L_2$ norm by $\|f\|_2=\{\E[f^2(\traj)]\}^{1/2}$. 
%
%
%

\subsection{Natural Stochastic Policies}

In OPE, $\epol$ is often pre-specified. Our focus is instead the case where $\epol$ depends on the natural value of the treatment in an observational study. Importantly, in this setting, both $\epol$ and $\bpol$ are \emph{unknown}. 
Natural stochastic policies are widely studied in the non-sequential (bandit) setting where $H=1$ \citep{MunozIvanDiaz2012PICE,HaneuseS.2013Eote}. However, it has not been extensively studied in the longitudinal (RL) setting but for a few exceptions. \citet{KennedyEdwardH2019NCEB} considers OPE with binary actions under a tilting policy in an NMDP (\emph{non}-Markov decision process). In comparison, we allow actions to be arbitrary and focus on the Markovian setting that is central to RL. \citet{YoungJessicaG2014Ieaa} considers OPE under a modified treatment policy in an NMDP using a parametric approach. In comparison, our methods are nonparametric and globally efficient, and we focus on the Markovian setting common in RL.

In this paper, we consider two types of natural stochastic policies: modified treatment policies and tilting polices.
These constructions are inspired by previous work focusing on the bandit and NMDP settings 
\citep{DiazIvan2020Cmaf,MunozIvanDiaz2012PICE,HaneuseS.2013Eote}.

\begin{definition}[Tilting policy]\label{def:tilting}
A tilting policy is specified by $u_t:\mathcal A_t\to\mathbb R$ and defined as
\begin{align}
\label{eq:tilting}
   \epol_t(a_t\mid s_t)=u_t(a_t)\bpol_t(a_t\mid s_t)\big/\int u_t(\tilde a_t)\bpol_t(\tilde a_t\mid s_t)\rd \tilde a_t.
\end{align} 
\end{definition}

Tilting policies tilt the behavior policy slightly toward actions with higher values of $u_t$. For example, for binary action, letting $u_t(1)=\delta,\,u_t(0)=1$ yields
\begin{align}
\label{eq:incremental}
  \epol_t(a_t\mid s_t)=\mathrm{I}(a_t=1)\frac{\delta \bpol_t(1\mid s_t)}{1+(\delta-1) \bpol_t(1\mid s_t)}+\mathrm{I}(a_t=0)\frac{\delta^{-1}\bpol_t(0\mid s_t)}{1+(\delta^{-1}-1)\bpol_t(0\mid s_t)}, 
\end{align}
as considered by \cite{KennedyEdwardH2019NCEB} in the binary-action NMDP setting. For $\delta=1$ we get $\epol=\bpol$; as $\delta$ shrinks, we tilt toward action 0; and, as $\delta$ grows, we tilt toward action 1. The parameter $\delta$ directly controls the amount of overlap; specifically $\epol_t(a_t\mid s_t)/\bpol_t(a_t\mid s_t)\leq\max(\delta,\delta^{-1})$.
For the general case in \cref{def:tilting}, we have that $\epol_t(a_t\mid s_t)/\bpol_t(a_t\mid s_t)\leq\max_{\tilde a_t}u_t(\tilde a_t)/\min_{\tilde a_t}u_t(\tilde a_t)$ so that the variation in $u_t$ can directly control the overlap. Tilting policies ensure that $\epol_t(\cdot\mid s_t)$ is absolutely continuous w.r.t. $\bpol_t(\cdot\mid s_t)$ so that the density ratio \emph{always} exists. In contrast, if $\epol_t$ is pre-specified and $\bpol_t$ is unknown, we cannot always ensure that the density ratio exists, let alone is bounded.

\begin{definition}[Modified treatment policy]\label{def:modified}
A modified treatment policy is specified by the maps $\tau_t:\mathcal S_t\times\mathcal A_t\to\mathcal A_t$ and assigns the action $\tau_t(s_t,a_t)$ in state $s_t$ when the natural action value is $a_t$. 
\end{definition}

Notice that the modified treatment policy is the same value $J$ as the evaluation policy defined by letting $\epol_t(\cdot\mid s_t)$ be the distribution of $a_t=\tau_t(s_t,\tilde a_t)$ under $\tilde a_t\sim\bpol_t(\cdot\mid s_t)$. For the purpose of OPE,
we can therefore equivalently define the modified treatment policy as this transformation of $\bpol$.
For example, if for each $s_t$, $\tau_t(s_t,\cdot)$ has a differentiable inverse $\tilde\tau_t(s_t,\cdot)$, then $\epol_t(a_t\mid s_t)=\bpol_t(\tilde\tau_t(s_t,a_t)\mid s_t)\tilde\tau'_t(s_t,a_t)$, where $'$ denotes a differentiation w.r.t $a_t$. The simplest example of a modified treatment policy is $\tau_t(s_t,a_t)=a_t+b_t(s_t)$ for some function $b_t(s_t)$, for which $\epol_t(a_t|s_t)=\bpol_t(a_t-b_t(s_t)\mid s_t)$. The function $b_t(s_t)$ quantifies the deviation from the natural value. Keeping $b_t(s_t)$ small ensures implementability.

\subsection{Off-Policy Evaluation}

Step-wise importance sampling \citep[IS;][]{precup2000eligibility} and direct estimation of $q$-functions \citep[DM;][]{munos2008finite,ernst2005tree} are two common approaches for OPE. However, the former is known to suffer from the high variance and the latter from model misspecification. To alleviate this, the doubly robust estimate combines the two \citep{MurphySA2001MMMf,jiang2016,thomas2016}. However, the asymptotic MSE of these can still grow exponentially in the horizon. \citet{KallusUehara2019} show that the efficiency bound in the TMDP case is actually \emph{polynomial} in $H$, $\bigO(CC'R^2_{\max}H^2/n)$, and give an estimator achieving it by combining marginalized IS \citep{XieTengyang2019OOEf} and $q$-modeling using cross-fold estimation. 
{When we additionally assume time invariance on TMDPs, \citet{KallusNathan2019EBtC} show an orders-smaller efficiency bound and develop an efficient estimator leveraging time invariance.}

All of the above methods focus on the case where $\epol$ is given explicitly. \emph{If} the behavior policy is known, then natural stochastic policies can also be regarded as given explicitly and these still apply. When $\bpol$ is unknown, as in observational studies, we can still operationalize these methods for evaluating natural stochastic policies by first estimating $\bpol$ from the data, plugging this into $\epol$
, and then treating $\epol$ as specified by this estimate. However, this will fail to be efficient,
as we discuss
in \cref{sec:tmdp}. In fact, the efficiency bounds for evaluating natural stochastic policies are \emph{different} than the pre-specified case.

\section{Efficiency Bounds}\label{sec:bounds}

In this section we calculate the efficiency bounds for evaluating natural stochastic policies in RL. We first briefly explain what the efficiency bound is (see \citealp{LaanMarkJ.vanDer2003UMfC} for more detail). 

A fundamental question is what is the smallest-possible error we can hope to achieve in estimating $J$. In parametric models, the \Cramer-Rao bound lower bounds the asymptotic MSE of all (regular) estimators. Our model, however, is nonparametric as it consists of \emph{all} TMDP distributions, \ie, \emph{any} choice for $p_t(r_t\mid s_t,a_t)$, $p_{t+1}(s_{t+1}\mid s_t,a_t)$, and $\pi_t(a_t\mid s_t)$ in \cref{eq:trajdist_mdp}.
Semiparametric theory gives an answer to this question 
by extending the notion of a \Cramer-Rao lower bound to nonparametric models. 
We first informally state the key property of the \emph{efficient influence function} (EIF) from semiparametric theory in our setting, i.e., the estimand is $J$ and the model is all TMDP distributions {denoted by $\cm_{\mathrm{TMDP}}$}.

\begin{theorem}[Informal description of Theorem 25.20 of \citet{VaartA.W.vander1998As}]\label{eq:vandervaartthm}
{The EIF $\phi(\traj)$ is the gradient of $J$ regarding $\cm_{\mathrm{TMDP}}$ that has the smallest $L_2$ norm, and it satisfies that, for any estimator $\hat J$ of $J$ that is regular w.r.t. $\cm_{\mathrm{TMDP}}$, $
\operatorname{AMSE}[\hat J]\geq \var[\phi(\traj)]$,
where $\operatorname{AMSE}[\hat J]$ is the second moment of the limiting distribution of $\sqrt{n}(\hat J-J)$.}

\end{theorem}
A regular estimator is any whose limiting distribution is insensitive to perturbations of order $\bigO(1/\sqrt{n})$ to the data-generating process $p_{\pi^b}$ that remain in $\cm_{\mathrm{TMDP}}$ (\ie, keep it a TMDP distribution). That is, a regular estimator is one that ``works'' for all problem instances in the model, and in this sense the result crucially depends on the model being considered.
Here, $\var[\phi(\traj)]$ is called the \emph{efficiency bound} as it is a lower bound on the asymptotic MSE of {all} regular estimators, which is a very general class.
This class is so general that in fact this also implies local minimax bound for \emph{all} estimators \citep[see][Theorem 25.21]{VaartA.W.vander1998As}. 


\subsection{Tilting Policies}

In the next result we calculate the EIF and efficiency bound for OPE of tiling policies. 
\begin{theorem}
\label{thm:mdp_example1}
Let $\epol$ be as in \cref{def:tilting}.
Then the EIF and efficiency bound of $J$ w.r.t. the model $\cm_{\mathrm{TMDP}}$ are, respectively,
\begin{align*}
    -J+\sum_{t=1}^{H}(\mu_t(r_t-v_t )+\mu_{t-1}v_t) ,\quad
    \Upsilon_{\TI1}=\sum_{t=0}^{H}\E[\var[\mu_t(r_t+v_{t+1})\mid s_t]],
\end{align*}
where $\mu_0=1,v_0=r_0=0$. Moreover, $\Upsilon_{\TI1}$ is upper bounded by $CC'R^2_{\max}H^2$. 
\end{theorem}

Note that the function $u_t$ that specifies the tilting policy is implicit in the variables $\mu_t,v_t$ above, which depend on $\epol_t$. While the efficiency bound is larger than in the case of a pre-specified evaluation policy \citep{KallusUehara2019}, the overall order, $CC'R^2_{\max}H^2$, is the same. {This implies we can circumvent the curse of horizon by using and efficient estimator.}

\begin{remark}[Comparison to pre-specified evaluation policy]
In a pre-specified evaluation policy case, \citet{KallusUehara2019} show that the EIF and efficiency bound are, respectively,
\begin{align}\label{eq:fixedpieeif}
    -J+\sum_{t=1}^{H}(\mu_t(r_t-q_t)+\mu_{t-1}v_t),\quad \sum_{t=0}^{H}\E[\mu^2_t\var[r_t+v_{t+1}\mid {s_t},a_t]]. 
\end{align}
Specifically, if we let $\epol$ be as in \cref{def:tilting} and assume that $\bpol$ is \emph{known} then this is the efficiency bound.
Compared with this quantity, $\Upsilon_{\TI1}$ is larger by $ \sum_{t=1}^{H} \E[\var[\mu_t q_t\mid s_t] ]$. 
\end{remark}

\begin{remark}[Non-Markovian Decision Processes]
\citet{KennedyEdwardH2019NCEB} provides the EIF for the binary-action tilting policy \cref{eq:incremental} under an NMDP. In comparison, our \cref{thm:mdp_example1} handles the Markovian case relevant to RL as well as a general action space. In particular, as we will see in \cref{sec:nmdp}, OPE under the NMDP model \textit{necessarily} suffers from the curse of horizon and therefore cannot handle long horizons.
An NMDP model can actually be embedded in a TMDP model (but \emph{not} vice versa) by including the whole state-action history up to time $t$ in the state variable $s_t$. Using this transformation, \cref{thm:mdp_example1} recovers the result of \citet{KennedyEdwardH2019NCEB} as a special case.
For further discussion of \citet{KennedyEdwardH2019NCEB}, refer to \cref{sec:nmdp}. 
\end{remark}

\begin{remark}[Bandit case]\label{rem:bandit_example1}
When $H=1$, 
the EIF is  $\eta_1(r_1-v_1(s_1))+v_1(s_1)$.
%
\end{remark}

\subsection{Modified Treatment Policies}

We next handle the case of modified treatment policies. {Again, we will see that we can potentially circumvent the curse of horizon with an efficient estimator.}

\begin{theorem}
\label{thm:mdp_example2}
Let $\epol$ be as in \cref{def:modified}.
Then the EIF and efficiency bound of $J$ w.r.t. the model $\cm_{\mathrm{TMDP}}$ are, respectively,
\begin{align}
-J+\sum_{t=1}^H \mu_t(r_t-q_t)+\mu_{t-1}q^{\tau}_t,\quad\Upsilon_{\MO1}=\sum_{t=0}^{H}\E[\mu^2_t\var[r_t+q^{\tau}_{t+1}\mid s_t,a_t]] 
\end{align}
where $q^{\tau}_t(s_t,a_t)=q_t(s_t,\tau_t(s_t,a_t))$. 
Moreover, $\Upsilon_{\MO1}$ is upper bounded by $CC'R^2_{\max}H^2$. 
\end{theorem}

\begin{remark}[Comparison to pre-specified evaluation policy]
Compared with the efficiency bound for a pre-specified evaluation policy, $\Upsilon_{\MO1}$ is larger by
$        \sum_{t=0}^{H} \E[ \mu^2_t\var[q^{\tau}_{t+1}\mid s_{t+1}] ].
$\end{remark}

\begin{remark}[Bandit case]\label{rem:bandit_example2}
When $H=1$, the EIF is $\eta_1(r_1-q_1(s_1,a_1))+q^{\tau}_1(s_1,a_1)$. This matches the results in \citet{DiazIvan2013Atce,Diaz2018}. 
\end{remark}

\section{Efficient and (Partially) Doubly Robust Estimation}\label{sec:tmdp}

We next propose efficient estimators for evaluating natural stochastic policies based on the obtained EIFs. 
Since both EIFs have second-order bias w.r.t. nuisances,
we can obtain efficient estimators by estimating nuisances under nonparametric rate conditions and plugging these into the EIFs with a sample-splitting cross-fitting technique \citep{ZhengWenjing2011CTME,ChernozhukovVictor2018Dmlf}.   

\subsection{Tilting Policies}

We propose an estimator $\hat J_{\TI1}$ for tilting polices in \cref{alg:tilting}. This is a meta-algorithm given estimation procedures for the nuisances $ w_t, \pi^b_t, q_t$, which we discuss how to estimate in \cref{sec:nuisance}. We next prove $\hat J_{\TI1}$ is efficient under nonparametric rate conditions on nuisance estimators, which crucially can be \emph{slower} than $\bigO_p(n^{-1/2})$ and do not require metric entropy conditions. 

\begin{algorithm}[t!]
 \caption{Efficient Off-Policy Evaluation for Natural Stochastic Policies}
 \begin{algorithmic}
 \label{alg:tilting}
\STATE Take a $K$-fold random partition of  $\{1,\dots,n\}=I_1\cup\cdots\cup I_K$ such that the size of each fold, $\abs{I_k}$, is within $1$ of $n/K$; set $\mathcal{U}_k=\{\traj^{(i)}:i\in I_k\},\,\mathcal{L}_k=\{\traj^{(i)}:i\notin I_k\}$
 \FOR{$k\in\{1,\cdots,K\}$}
    \STATE Using only $\mathcal{L}_k$ as data, construct nuisance estimators ${\hat w^{(k)}}_t,\,{\hat \pi^{b,(k)}}_t,{\hat q^{(k)}}_t$ for $t\leq H$
    \STATE Set $\hat\pi^{e,(k)}_t(a_t\mid s_t)=u_t(a_t)\hat \pi^{b,(k)}_t(a_t\mid s_t)/\int u_t(\tilde a_t)\hat \pi^{b,(k)}_t(\tilde a_t\mid s_t)\rd \tilde a_t$
    \STATE \phantom{Set} $\hat \eta^{(k)}_t(s_t,a_t)=\hat\pi^{e,(k)}_t(a_t\mid s_t)/\hat \pi^{b,(k)}_t(a_t\mid s_t),\,\hat v^{(k)}_t(s_t)=\int \hat q^{(k)}_t(s_t,a_t)\hat \pi^{e,(k)}_t(a_t\mid s_t)\rd a_t$
    \STATE Set $\hat{J}_k=\frac1{|I_k|}\sum_{\traj\in\cu_k}\hat\phi^{(k)}(\traj)$, where 
    \begin{align}\hspace{-2.225em}
    \label{eq:central}
 \hat\phi^{(k)}(\traj)=\sum_{t=1}^H \hat w^{(k)}_t(s_t)\hat\eta^{(k)}_t(s_t,a_t)(r_t-\hat v^{(k)}_t(s_t))+\hat w^{(k)}_{t-1}(s_{t-1})\hat\eta^{(k)}_{t-1}(s_{t-1},a_{t-1})\hat v^{(k)}_{t}(s_{t})
\end{align}
    \ENDFOR 
    \STATE Return
    $\hat J_{\TI1} = \frac{1}{n}\sum^K_{k=1}|I_k|\hat{J}_k$
\end{algorithmic}
\end{algorithm}

\begin{theorem}[Efficiency] \label{thm:efficiency_tmdp1}
Suppose $\forall k\leq K,\forall j \leq H$,
$\|\hat w^{(k)}_j(s_j)-w_j(s_j)\|_2\leq\alpha_1$, $\|\hat \pi^{b,(k)}_j(a_j\mid s_j)-\pi^b_j(a_j\mid s_j)\|_2\leq\alpha_2$, $\|\hat q^{(k)}_j(s_j,a_j)-q_j(s_j,a_j)\|_2\leq\beta$, where $\alpha_1=\bigO_p(n^{-1/4})$, $\alpha_2=\op(n^{-1/4})$, $\beta=\bigO_p(n^{-1/4})$, $\alpha_1\beta=\op(n^{-1/2})$. Then, $\sqrt{n}( \hat J_{\TI1}-{J})\stackrel{d}{\longrightarrow} \bN(0,\Upsilon_{\TI1})$. 
\end{theorem}
The result essentially follows by showing that $|{\hat J_{\TI1}-J-\P_n[\phi(\traj)]}|\leq\alpha_1\alpha_2+\alpha_1\beta+\alpha_2\beta+\alpha^2_2+o_p(n^{-1/2})$, where $\phi(\traj)$ is the EIF. Under the above rate assumptions, the right-hand side is $\op(n^{-1/2})$ and the result is immediately concluded from CLT.
Notice that \emph{if} we knew the behavior policy so that $\alpha_2=0$, this becomes simply $\alpha_1\beta+o_p(n^{-1/2})$ and we recover the doubly robust structure of the pre-specified case \citep{KallusUehara2019}: the estimator is consistent if \emph{either} $w_t$ \emph{or} $q_t$ is consistently estimated.
In our setting, because of the term $\alpha^2_2$, the consistent estimation of $\bpol$ is always required to estimate $J$ consistently. So, we have a \emph{partial} double robustness in the sense that the estimator is consistent as long as $\bpol$ \emph{and} either $w$ or $q$ are consistently estimated. 

\begin{theorem}[Partial double robustness] \label{thm:partialdr_tmdp1} 
Suppose $\forall k\leq K,\forall j \leq H$, for some $w^{\dagger}_j,q^{\dagger}_j$, $\|\hat w^{(k)}_j(s_j)-w^{\dagger}_j(s_j)\|_2=\op(1)$, $\|\hat q^{(k)}_j(s_j,a_j)-q^{\dagger}_j(s_j,a_j)\|_2=\op(1)$, and $\|\hat \pi^{b,(k)}_j(a_j|s_j)-\pi^b_j(a_j|s_j)\|_2=\op(1)$. As long as either $q^{\dagger}_j=q_j$ or $w^{\dagger}_j=w_j$, we have $\hat J_{\TI1}\stackrel{p}{\longrightarrow} J$. 
\end{theorem}

\begin{remark}[Comparison to \citealp{KallusUehara2019}]\label{rem:drlcomp}
Since we \emph{have to} estimate $\bpol$ (and hence $\epol$) consistently anyway for our estimator to work,
a careful reader might wonder whether we might as well plug in the estimated $\epol$ into estimators that are efficient for the  pre-specified case such as \citet{KallusUehara2019}. Specifically, we could replace \cref{eq:central} in \cref{alg:tilting} with
\begin{align*}
   \phi^{(k)}(\mathcal T)=\sum_{t=1}^{H} \hat w^{(k)}_t(s_t)\hat\eta^{(k)}_t(s_t,a_t)(r_t-\hat q^{(k)}_t(s_t,a_t))+  \hat w^{(k)}_{t-1}(s_{t-1})\hat\eta^{(k)}_{t-1}(s_{t-1},a_{t-1})\hat v^{(k)}_t(s_t),
\end{align*}
which corresponds to plugging our estimated nuisances into the EIF derived in \citet{KallusUehara2019}.
However, this can fail to achieve a $\sqrt{n}$-convergence rate, let alone efficiency. Specifically, in \cref{thm:efficiency_tmdp1}, we used the fact that \cref{eq:central} has a second-order bias structure w.r.t. $w_t,\pi^b_t,q_t$ to ensure the $\sqrt{n}$-consistency and efficiency. In contrast, the above does {not} have this structure. It only has such a structure when $\hat v^{(k)}_t$ is the integral of $\hat q^{(k)}_t$ with respect to the \emph{true} $\epol_t$, which \cite{KallusUehara2019} use to achieve efficiency, but that is not the case here. 
\end{remark}

\begin{remark}[Estimation of $v$-functions]
Although $\hat q^{(k)}_t$ does not explicitly appear in \cref{eq:central}, we do need to estimate $\hat q^{(k)}_t$ first and then compute $\hat v^{(k)}_t$ based on it as in \cref{alg:tilting}, instead of directly estimating $v_t$. The reason is that we cannot generally say that \cref{eq:central} has a second-order bias structure w.r.t. $w_t,\bpol_t,v_t$. Therefore, the efficiency result would not be guaranteed. 
To achieve the efficiency, it is crucial to use the specific construction of $\hat v^{(k)}_t$ in \cref{alg:tilting}, which ensures a certain compatibility between the nuisance estimators, as they all use the same estimated behavior policy.
\end{remark}

\subsection{Modified Treatment Policies}

We similarly define the estimator $\hat J_{\MO1}$ for the case of modified treatment policies by taking \cref{alg:tilting} and (a) replacing $\hat \pi^{e,(k)}_t(a_t\mid s_t)$ by $\hat \pi^{e,(k)}_t(a_t\mid s_t)=\hat\pi^{b,(k)}_t(\tilde\tau_t(s_t,a_t)\mid s_t)\tilde\tau'_t(s_t,a_t)$ and (b) replacing \cref{eq:central} by
\begin{align*}\hat\phi^{(k)}(\traj)=
\sum_{t=1}^H \hat w^{(k)}_t(s_t)\hat\eta^{(k)}_t(s_t,a_t)(r_t-\hat q^{(k)}_t(s_t,a_t))+\hat w^{(k)}_{t-1}(s_{t-1})\hat\eta^{(k)}_{t-1}(s_{t-1},a_{t-1})\hat q^{(k)}_t(s_t,\tau_t(s_t,a_t)). 
\end{align*}
We then have the following efficiency and (full) double robustness results.  
\begin{theorem}[Efficiency]\label{thm:efficiency_tmdp2}
Suppose $\forall k\leq K,\forall j \leq H$, $\|\hat w^{(k)}_j(s_j)-w_j(s_j)\|_2\leq\alpha_1,\|\hat \pi^{b,(k)}_j(a_j|s_j)- \pi^{b,(k)}_j(a_j|s_j)\|_2\leq\alpha_2,\|\hat q^{(k)}_j(s_j,a_j)-q_j(s_j,a_j)\|_2\leq\beta$ where $(\alpha_1+\alpha_2)\beta=\op(n^{-1/2}),\,\max\{\alpha_1,\alpha_2,\beta\}=\op(1)$. Then, $\sqrt{n}( \hat J_{\MO1}-{J})\stackrel{d}{\longrightarrow} \bN(0,\Upsilon_{\MO1})$. 
\end{theorem}
\begin{theorem}[Double robustness]\label{thm:dr_estimatino2_tmdp}
Assume  $\forall k\leq K,\forall j \leq H$, for some $\pi^{b\dagger}_j,q^{\dagger}_j,w^{\dagger}_j$, $\|w^{(k)}_j(s_j)-w^{\dagger}_j(s_j)\|_2=\op(1),\,\|\hat \pi^{b}_j(a_j|s_j)- \pi^{b\dagger}_j(a_j|s_j)\|_2=\op(1),\,\|\hat q^{(k)}_j(s_j,a_j)-q^{\dagger}_j(s_j,a_j)\|_2=\op(1)$. Then as long as either $q^{\dagger}_j=q_j$ or $\pi^{b\dagger}_j=\pi^b_j,w^{\dagger}_j=w_j$, we have $\hat J_{\MO1}\stackrel{p}{\longrightarrow} J$. 
\end{theorem}
These theorems arise from the bias structure $|\hat J_{\MO1}-J-\P_n[\phi(\traj)]|\leq(\alpha_1+\alpha_2)\beta+o_p(n^{-1/2})$. The conditions on nuisance estimates in these theorems are weaker than the ones for tilting policies. Comparing \cref{thm:efficiency_tmdp1,thm:efficiency_tmdp2}, the condition in \cref{thm:efficiency_tmdp2} is satisfied even if some of $\alpha_1,\alpha_2,\beta$ are \emph{slower} than $\op(n^{-1/4})$. Comparing \cref{thm:dr_estimatino2_tmdp,thm:partialdr_tmdp1}, the condition in \cref{thm:dr_estimatino2_tmdp} is satisfied even if the behavior policy model is misspecified. The intuitive reason is that for a modified treatment policy, $J$
can be specified in a form not depending on $\bpol$, while this is not true for tilting policies.

\subsection{Nuisance Estimation}\label{sec:nuisance}

Our estimators for both types of stochastic policies require that we estimate $\bpol_t,w_t,q_t$ at some slow rate.
Here we discuss some standard ways to estimate these nuisance functions. We focus on the case of tilting policies for brevity. 

{First of all, estimating $\bpol_t$ amounts to fitting a probabilistic classification model in the case of finitely many actions and to conditional density estimation in the case of continuous actions. Once we fit $\bpol_t$, we also immediately have an estimate of $\epol_t$. We can then use standard methods for estimating $w_t$ and $q_t$ that assume $\epol_t$ is given by plugging in our estimate for it as follows.  Generally speaking, if the estimate for $q_t$ or $w_t$ would have had some convergence rate $r_n$ if $\epol_t$ were given exactly, then this rate does not deteriorate as long as the plugged-in estimate for $\epol_t$ also has rate $r_n$. For all of our nuisance estimators, we do not require any metric entropy conditions.} 

\paragraph{Estimation of $q$-functions}

In the tabular case, a model-based approach is the most common way to estimate $q$-functions from off-policy data. In the non-tabular case, we have to rely on some function approximation. The key equation to derive these methods is the Bellman equation:
\begin{align*}
    q_t(s_t,a_t)=\E[r_t+q_{t+1}(s_{t+1},\epol)\mid s_t,a_t]. 
\end{align*}
where $q_{t}(s_{t},\pi)=\int q_t(s_t,a_t)\pi(a_t\mid s_t)\mathrm{d}a_t$. One of the most common ways to operationalize this is using fitted $q$-iteration (\citealp{antos2008learning,LeHoang2019BPLu,DuanYaqi2020MOEw}; expressed here using an estimated evaluation policy, $\hat \pi ^e$):
    set $\hat q_{H+1}\equiv0$, and 
    for $t=H,\dots,1$
        estimate $\hat q_t$ by regressing $r_t+\hat q_{t+1}(s_{t+1},\hat \pi^e)$ onto $s_t,a_t$ using any given regression method.
Note that the regressions can also be replaced with $Z$-estimation \citep{Ueno2011,LuckettDanielJ.2018EDTR} based on the moment equations:
\begin{align*}
    \E[g_t(s_t,a_t)\prns{r_t+q_{t+1}(s_{t+1},\epol)-q_t(s_t,a_t)}]=0,&&\forall g_t,\,t\leq H. 
\end{align*}
When we let $q_t$ and $g_t$ be the class of linear functions, this leads to the LSTD method \citep{LagoudakisMichail2004LPI}.

\paragraph{Estimation of Marginal Density Ratios}

In the tabular case, a model-based approach \citep{Yin2020} would be a competitive way to estimate marginal density ratios:
\begin{align*}
 \hat  w_t(s_t)=\frac1{\hat p_{\bpol_t}(s_t)}\int \hat p_{t}(s_{t}|s_{t-1},a_{t-1})\prod_{k=0}^{t-1}\prns{ \epol_k(a_k|s_k)\hat p_{k}(s_{k}|s_{k-1},a_{k-1})}\mathrm{d}(h_{a_{t-1}}), 
\end{align*}
where $\hat p_{{t}},\hat p_{\bpol_t}$ are each empirical frequency (histogram) estimators, and $h_{a_{t-1}}=(s_1,a_1,\cdots,a_{t-1})$. When the behavior policy is known, \citet{XieTengyang2019OOEf} also proposed a similar method to estimate $w_t$. In the non-tabular case, we have to rely on some function approximation methods. In this case, we can use the equation $w_t=\E[\lambda_{t-1}\mid s_t]$. Thus, for example, $w_t$ is estimated by regressing $\lambda_{t-1}$ onto $s_t$. Alternatively, we can use the recurrence $w_t=\E[\eta_{t-1}w_{t-1}\mid s_t]$ and recursively construct regression estimates similarly to fitted $q$-iteration but going forward in time rather than backward. 

\section{Application to Non-Markov Decision Processes}\label{sec:nmdp}

So far, we have so focused on the TMDP setting where the trajectory distribution $p_\pi$ is given by \cref{eq:trajdist_mdp}. For the sake of completeness and comparison, we next also consider the \emph{non-Markov} decision process (NMDP) model.
Our results for TMDP may in fact be applied directly 
by recognoizing that an NMDP can be embedded in a TMDP by including the whole history in the state variable. (The opposite cannot be done and the results for TMDP are novel.)
We use this to derive the efficiency bound for the NMDP case and compare to \citet{KennedyEdwardH2019NCEB,DiazIvan2020Nceb}.

Relaxing the Markov assumption of the TMDP model, yields the NMDP model $\cm_{\mathrm{NMDP}}$, where the trajectory distribution $p_{\pi}(\traj)$ is
\begin{align} \notag
p_1(s_1)\prod_{t=1}^H\pi_t(a_t\mid \cjj_{s_t})p_t(r_t\mid \cjj_{a_t})p_{t+1}(s_{t+1}\mid \cjj_{r_t}), 
\end{align}
where $\cjj_{a_t}$ is $(s_1,a_1,r_1,\cdots,a_t)$, $\cjj_{s_t}$ is $(s_1,a_1,r_1,\cdots,s_t)$ and  $\cjj_{r_t}$ is $(s_1,a_1,r_1,\cdots,r_t)$. 

Notice that in the embedding of an NMDP as a TMDP, the marginal density ratio $\mu_t$ in the TMDP is \textit{exactly} the cumulative importance ratio $\lambda_t$ in the original NMDP. We therefore have the following for tilting policies as a corollary of \cref{thm:mdp_example1}.
\begin{theorem}
\label{thm:nmdp_example1}
Let $\epol$ be as in \cref{def:tilting}.
Then the EIF and the efficiency of $J$ bound w.r.t. the model $\cm_{\mathrm{NMDP}}$ are, respectively,
\begin{align*}
\sum_{t=1}^{H} \lambda_t\{r_t-v_t(\cjj_{s_t})\}+\lambda_{t-1}v_t(\cjj_{s_t}),\,\sum_{t=0}^{H}\E[\lambda^2_{t-1}(\cj_{s_{t-1}})\var[\eta_t(\cj_{a_t})\{R_t+v_{t+1}(\cj_{s_{t+1}})\}\mid \cj_{s_t}]].
\end{align*}
\end{theorem}
When actions are binary ($\mathcal A_t=\{0,1\}$) and $r_1=\cdots=r_{H-1}=0$ so rewards only occur at the end ($r_H$), this coincides with Theorem 2 of \citet{KennedyEdwardH2019NCEB}.

The main issue with the NMDP model is that OPE estimators that are regular w.r.t. the NMDP model necessarily suffer from the curse of horizon, in the sense that their MSE is exponentially growing in horizon.
\begin{theorem}\label{thm:nmdp_variance}
Suppose $\exp(\E_{p_{\epol}}[\log(\eta_t)])\geq C_{\min}$ and $\exp(\E[\log(\var[\eta_t(\cjj_{a_t})\{r_t+v_{t+1}(\cjj_{s_{t+1}})\}\mid \cjj_{s_t}])])\geq V_{\min}^2$. Then, the efficiency bound in \cref{thm:nmdp_example1} is lower bounded by $C_{\min}^{H-1}V_{\min}^2$.
\end{theorem}

Notice $\E_{p_{\epol}}[\log(\eta_t)]$ is exactly the Kullback-Leibler divergence between $\epol(\cdot\mid s_t)$ and $\bpol(\cdot\mid s_t)$ averaged over $p_{\epol}(s_t)$.

We can similarly extend \cref{thm:mdp_example2} (modified treatment policies) to the NMDP case as follows, and again see that it suffers from the curse of horizon. 
\begin{theorem}
\label{thm:nmdp_example2}
Let $\epol$ be as in \cref{def:modified}.
Then the EIF and efficiency bound of  $J$  w.r.t. the model $\cm_{\mathrm{NMDP}}$ are, respectively,
\begin{align*}
    \sum_{t=1}^H \lambda_t\{r_t-q_t(
    \cjj_{a_t})\}+\lambda_{t-1}q_t(\cjj_{s_t},\tau_t(\cjj_{a_t})),\,\sum_{t=0}^{H}\E[\lambda^2_t \var[r_t+q_{t+1}(\cjj_{s_{t+1}},\tau_{t+1}(\cjj_{a_{t+1}}))\mid \cjj_{a_t}]].
\end{align*}
\end{theorem}
Again, when $r_1=\cdots=r_{H-1}=0$ so rewards only occur at the end ($r_H$), this coincides with Theorem 2 of \citet{DiazIvan2020Nceb}.
Again, this still suffers from the curse of horizon.

\begin{theorem}\label{thm:nmdp_variance2}
Suppose $\exp(\E_{p_{\epol}}[\log(\eta_t)])\geq C_{\min}$ and $\exp(\E[\log(\var[r_t+q_{t+1}(\cjj_{s_{t+1}},\tau_{t+1}(\cjj_{a_{t+1}}))\mid \cjj_{a_t}])])\geq V_{\min}^2$. Then, the efficiency bound in \cref{thm:nmdp_example2} is lower bounded by $C_{\min}^{H}V_{\min}^2$.
\end{theorem}

\section{Extension to Infinite-Horizon Time-Homogeneous Markov Decision Processes}\label{sec:mdp}

We next extend the results to time-homogeneous MDPs (MDPs) where transitions, rewards, and policies do not depend on $t$, i.e., $p_{t}(r|s,a)=p(r|s,a),p_{t}(s'|s,a)=p(s'|s,a),\bpol_t=\bpol,\tau_t=\tau$. Here, the estimand we consider is an average discounted reward, $J(\gamma)=(1-\gamma)\lim_{T\to \infty} \E_{\epol}[\sum_{t=1}^T \gamma^{t-1} r_t]$ when the initial state distribution is $p_e^{(1)}(s)$. {In this section, We derive the EIF and efficiency bound of $J(\gamma)$ w.r.t. the  MDP model. Then, we discuss the relationship to the efficiency bound and estimators for pre-specified evaluation policies.}

Although we can still apply methods for TMDP in the MDP case, if we correctly leverage the time-homogeneity of MDP we should do much better in that the rate of MSE should be $\bigO(1/NH)$, not $\bigO(1/N)$, when we observe $N$ trajectories of length $H$. Per \citet{Liu2018},
\begin{equation}\label{eq:infhorizonJ} J(\gamma)=
\E_{s\sim p_{b}(s),a\sim \bpol(a|s),r\sim p(r|s,a)}[r\;{p^{(\infty)}_{e,\gamma}(s)\epol(a\mid s)}/({p_{b}(s)\bpol(a\mid s)})],\end{equation}
where $p_{b}(s)$ is any distribution\footnote{{In practice, it is often take as the stationary distribution associated with the MDP and the behavior policy $\bpol$. Our theory holds not only for this case but also for any $p_b(s)$.}} and $p^{(\infty)}_{e,\gamma}$ is the $\gamma$-discounted average visitation distribution associated with the MDP, $\epol$, and the initial state distribution $p_{e}^{(1)}(s)$. 

When $\epol$ is pre-specified, \cite{KallusNathan2019EBtC} derived the efficiency bound of $J(\gamma)$ under a nonparametric model 
and proposed an efficient estimator. Here, we present corresponding results for the efficient evaluation of natural stochastic policies in time-homogeneous infinite-horizon RL. 
We consider the observed data to be $n$ i.i.d. draws from the stationary behavior distribution: for $i=1,\dots,n$,\footnote{Considering the data to instead be $N=n/H$ observations of $H$-long trajectories can be also handled if we impose some additional mixing assumptions following \citet{KallusNathan2019EBtC}, which ensures the dependent trajectories look approximately like independent observations. This only changes the analysis. The estimator itself is unchanged.}\footnote{The observation of the ensuing action $a'^{(i)}$ is actually only needed for the case of modified treatment policies.}
\begin{align*}
    (s^{(i)},a^{(i)},r^{(i)},s'^{(i)},a'^{(i)}) \sim p_{b}(s,a,r,s',a')=p_{b}(s)\bpol(a\mid s)p(s'|s,a)p(r|s,a)\bpol(a'\mid s'). 
\end{align*}
We consider a fully nonparametric model $\cm_{\mathrm{MDP}}$ in that we make no restrictions on the above distributions. In this section, we define $q(s,a)=\E_{p_{\pi_e}}[\sum_{t=1}^{\infty} \gamma^{t-1} r_t |s_1=s,a_1=a]$, $q^{\tau}(s,a)=q(s,\tau(s,a))$, $v(s)=\E_{\epol(a|s)}[q(s,a)|s]$, $w^{*}(s)=p^{(\infty)}_{e,\gamma}(s)/p_{b}(s)$, and $\mu^{*}(s,a)=w^{*}(s)\eta(s,a)$.

\subsection{Tilting Policies}

We consider first the case of tilting policies. 

\begin{theorem}
\label{thm:mdpmdp_example1}
Let $\epol$ be as in \cref{def:tilting}.
Then the EIF and efficiency bound of $J(\gamma)$ w.r.t. $\cm_{\mathrm{MDP}}$ are, respectively,
\begin{align*}
   \mu^{*}(s,a)(r+\gamma v(s')-v(s)),\quad\Upsilon_{\TI2}=\E[\var[\mu^{*}(s,a)(r+\gamma v(s'))\mid s]]. 
\end{align*}
\end{theorem}

We can construct an efficient estimator by following a similar but slightly different cross-fitting strategy as before. 
With additional data $s^{(j)}_{1}\sim p^{(1)}_{e}(s),\,j=1,\dots,m$ where $m=\Omega(n)$ (or, if $p^{(1)}_e$ is known), and given nuisance estimators $\hat \pi^{b,(k)},\hat q^{(k)},\hat w^{*(k)}$, we propose the estimator $\hat J_{\TI2}$ for $J(\gamma)$ by taking \cref{alg:tilting} and replacing $\hat J_k$ with  
\begin{align}
\label{eq:tilting_infinite}
   &\hat J_k=\frac{1-\gamma}{m}\sum_{j=1}^m\hat v^{(k)}(s^{(j)}_1)+\frac{1}{|I_k|}\sum_{i\in I_k}{\hat w^{*(k)}(s^{(i)})\hat \eta^{(k)}(s^{(i)},a^{(i)})(r^{(i)}+\gamma \hat v^{(k)}(s^{'(i)})-\hat v^{(k)}(s^{(i)}))},
   \\
   &\hat\pi^{e,(k)}(a\mid s)=u(a)\hat \pi^{b,(k)}(a\mid s)\big/\int u(\tilde a)\hat \pi^{b,(k)}(\tilde a\mid s)\rd \tilde a,\nonumber\\
   &\hat \eta^{(k)}(s,a)=\hat\pi^{e,(k)}(a\mid s)/\hat\pi^{b,(k)}(a\mid s),\,\hat v^{(k)}(s)=\int \hat q^{(k)}(s,a)\hat\pi^{e,(k)}(a\mid s)\mathrm{d}a.  \label{eq:tilting_infinite_etav}
\end{align}
To estimate $\bpol$, we can follow \cref{sec:nuisance}. To estimate $w^*, q$, we can solve the following conditional moment equations \citep[cf.][]{Liu2018,KallusNathan2019EBtC} either by instantiating with some set test functions and solving the empirical moments or using a method such as \citet{bennett2019deep}:
\begin{align} \label{eq:nuisance1}
    0 &=(1-\gamma)\E_{s_1\sim p_e^{(1)}}[f(s_1)]+\E_{(s,a,s')\sim p_b}[\gamma w^{*}(s)(\eta(s,a)f(s')-f(s))]&&\forall f(s),\\
    \label{eq:nuisance1b}
    0&=\E_{(s,a,r,s')\sim p_b}[g(s,a)(r+\gamma v(s')-q(s,a))]&&\forall g(s,a).
\end{align}

Given any nuisance estimators that satisfy certain weak rate assumptions (and no other assumptions),
we prove $\hat J_{\mathrm{TI2}}$ is efficient and partially doubly robust. 
\begin{theorem}[Efficiency]
\label{thm:mdp_estimation1}
Suppose $\forall k\leq K$, $\|\hat w^{*(k)}(s)-w^{*}(s)\|_2\leq\alpha_1,\|\hat \pi^b(a|s)-\pi^b(a|s) \|_2\leq\alpha_2, \|\hat q^{(k)}(s,a)-q(s,a)\|_2\leq\beta$, where $\alpha_1=\bigO_p(n^{-1/4}),\alpha_2=\op(n^{-1/4}),\beta=\bigO_p(n^{-1/4}),\alpha_1\beta=\op(n^{-1/2})$. Then, $\sqrt{n}( \hat J_{\TI2}-J)\stackrel{d}{\rightarrow}\cN(0,\Upsilon_{\TI2})$.
\end{theorem}

\begin{theorem}[Partial Double Robustness]
\label{thm:dr_estimatino1_mdp}
Assume $\forall k\leq K$, for some $w^{*\dagger}(s),q^{\dagger}(s,a)$, $\|\hat w^{*(k)}(s)-w^{*\dagger}(s)\|_2=\op(1),\|\hat \pi^b(a|s)-\pi^{b}(a|s) \|_2=\op(1), \|\hat q^{(k)}(s,a)-q^{\dagger}(s,a)\|_2=\op(1)$. Then, as long as $w^{*\dagger}(s)=w^{*}(s)$ or $q^{\dagger}(s,a)=q(s,a)$, $\hat J_{\TI2} \stackrel{p}{\rightarrow} J.$
\end{theorem}

The result again essentially follows by showing that $|{\hat J_{\TI1}-{J}-\P_n[\phi_{}(s,a,r,s')]}|\leq\alpha_1\alpha_2+\alpha_1\beta+\alpha_2\beta+\alpha^2_2+o_p(n^{-1/2})$, where $\phi_{}(s,a,r,s')$ is the EIF. Under the above rate assumptions, the right-hand side is $\op(n^{-1/2})$ and the result is immediately concluded from CLT. Partially doubly robustness similarly holds as in \cref{sec:tmdp}.

\subsection{Modified Treatment Policies}
We next consider the case of modified treatment policies. 

\begin{theorem}
\label{thm:mdpmdp_example2}
Let $\epol$ be as in \cref{def:modified}.
Then the EIF and efficiency bound of $J(\gamma)$ w.r.t. $\cm_{\mathrm{MDP}}$ are, respectively,
\begin{align*}
  \mu^{*}(s,a)(r+\gamma q^{\tau}(s',a')-q(s,a)),\quad\Upsilon_{\MO2}=\E[\mu^{*2}(s,a)\var[r+\gamma q^{\tau}(s',a')\mid s,a]]. 
\end{align*}
\end{theorem}
 
We can construct an efficient estimator as follows. With additional data $(s^{(j)},a^{(j)})\sim p^{(1)}_{e}(s)\bpol(a\mid s)$, $j=1,\dots,m$
where $m=\Omega(n)$, and given nuisance estimators $\hat w^{*(k)},\hat \pi^{b,(k)},\hat q^{(k)}$, we propose the estimator $\hat J_{\MO2}$
by taking \cref{alg:tilting} and replacing $\hat J_k$ with  
\begin{align*}
\hat J_{k}=&\frac1{|I_k|}\sum_{i\in I_k}\hat w^{*(k)}(s)\hat\eta^{(k)}(s^{(i)},a^{(i)})(r^{(i)}+\gamma \hat q^{(k)\tau}(s^{'(i)},a^{'(i)})-\hat q^{(k)}(s^{(i)},a^{(i)}))\\
& +\frac{(1-\gamma)}{m}\sum_{j=1}^m\hat q^{(k)\tau}(s^{(j)},a^{(j)}),\quad\hat \pi^{e,(k)}(a_t\mid s_t)=\hat\pi^{b,(k)}(\tilde\tau(s_t,a_t)\mid s_t)\tilde\tau'(s_t,a_t),
\end{align*}
and $\hat\eta^{(k)},\,\hat v^{(k)}$ are as in \cref{eq:tilting_infinite_etav}. To estimate $w^{*}$ we can use \cref{eq:nuisance1} and to estimate $q$ we can use: 
\begin{align} \label{eq:nuisance2}
  0 = 
  \E_{(s,a,r,s',a')\sim p_b}[g(s,a)\prns{r-q(s,a)+\gamma q^{\tau}(s',a')}]&&\forall g(s,a).
\end{align}
We next prove $\hat J_{\MO2}$ is efficient and doubly robust. 

\begin{theorem}[Efficiency]
\label{thm:mdp_estimation2}
Assume $\forall k\leq K$, $\|\hat w^{*(k)}(s)-w^{*}(s)\|_2\leq\alpha_1,\|\hat \pi^{b,(k)}(a|s)-\pi^b(a|s)\|_2\leq\alpha_2,\|\hat     q^{(k)}(s,a)-q(s,a)\|_2\leq\beta$, where $(\alpha_1+\alpha_2)\beta=\op(n^{-1/4}),\max\{\alpha_1,\alpha_2,\beta\}=\op(1)$. Then, $\sqrt{n}(\hat J_{\MO2}-J)\stackrel{d}{\rightarrow}\cN(0, \Upsilon_{\MO2})$. 
\end{theorem}

\begin{theorem}[Double Robustness]
\label{thm:dr_estimatino2_mdp}
Assume $\forall k\leq K$, for some $w^{*\dagger},\pi^{b\dagger},q^{\dagger}$, $\|\hat w^{*(k)}(s)-w^{*\dagger}(s)\|_2=\op(1),\|\hat \pi^{b,(k)}(a|s)-\pi^{b\dagger}(a|s)\|_2=\op(1),\|\hat q^{(k)}(s,a)-q^{\dagger}(s,a)\|_2=\op(1)$. Then, as long as $w^{*\dagger}=w^{*},\pi^{b\dagger}=\bpol$ or $q^{\dagger}=q$, $\hat J_{\MO2}\stackrel{p}{\rightarrow}J$. 
\end{theorem}

\subsection{Comparison with \citet{KallusNathan2019EBtC,tang2019harnessing}}\label{sec:tang}

We next compare our results with some related work for the case of pre-specified policies.  

\paragraph{Comparison with  \citet{KallusNathan2019EBtC}}

When the evaluation policy {is} pre-specified, \citet{KallusNathan2019EBtC} proposed an estimator that is similar but uses $\hat q^{(k)}$ in place of the last $\hat v^{(k)}$ in \cref{eq:tilting_infinite}. Then, under similar rate conditions to \cref{thm:mdp_estimation1}, they prove it achieves the asymptotic MSE $$\Upsilon_{\mathrm{PR}}=\E[\mu^{*2}(s,a)\var[r+\gamma v(s')\mid s,a]],$$
which is the efficiency bound when the evaluation policy is pre-specified so the estimator is efficient.
Notice that $\Upsilon_{\mathrm{PR}}$ is smaller than $\Upsilon_{\TI2}$ by $\E[w^{*2}(s)\var[\eta(s,a)(r+\gamma v(s')) \mid s]]$. As in \cref{rem:drlcomp}, na\"ively plugging in an estimated $\epol$ into the EIF for the pre-specified case can fail to be efficient or even $\sqrt n$-consistent. 

\paragraph{Comparison with \citet{tang2019harnessing}}

In the case of a pre-specified evaluation policy and \textit{known} behavior policy, \citet{tang2019harnessing} propose an estimator with a form similar to \cref{eq:tilting_infinite} without sample splitting and where $\hat v$ is \textit{directly} estimated (rather than computed as a function of other nuisance estimates). The similarity to \cref{eq:tilting_infinite} appears to be coincidental and superficial. In the case of pre-specified evaluation policy,
even if we used oracle values for all nuisances, the estimator of \citet{tang2019harnessing} is \textit{inefficient} since its variance would be equal to $\Upsilon_{\TI2}$, which is \textit{larger} than $\Upsilon_{\mathrm{PR}}$.
\citet{tang2019harnessing} do not study the asymptotic properties of their estimator, but we can show that using cross-fitting of nuisance esimates, the asymptotic MSE of the estimator is also equal to $\Upsilon_{\TI2}$ (see \cref{sec:tangappendix}). Again, this is inefficient in the case of a pre-specified evaluation policy.
And, in the case of a natural stochastic policy, $\hat v$ must be computed in a fashion compatible with $\hat q$ and $\hat \bpol$ in order to ensure the second-order bias structure and hence efficiency.

\section{Empirical Study}\label{sec:empirical}

\begin{figure}[t!]\footnotesize%
\begin{minipage}{0.025\textwidth}
\rotatebox{90}{~~~Tilting Policy}
\end{minipage}%
\begin{minipage}{0.325\textwidth}
\centering
\includegraphics[width=.95\linewidth]{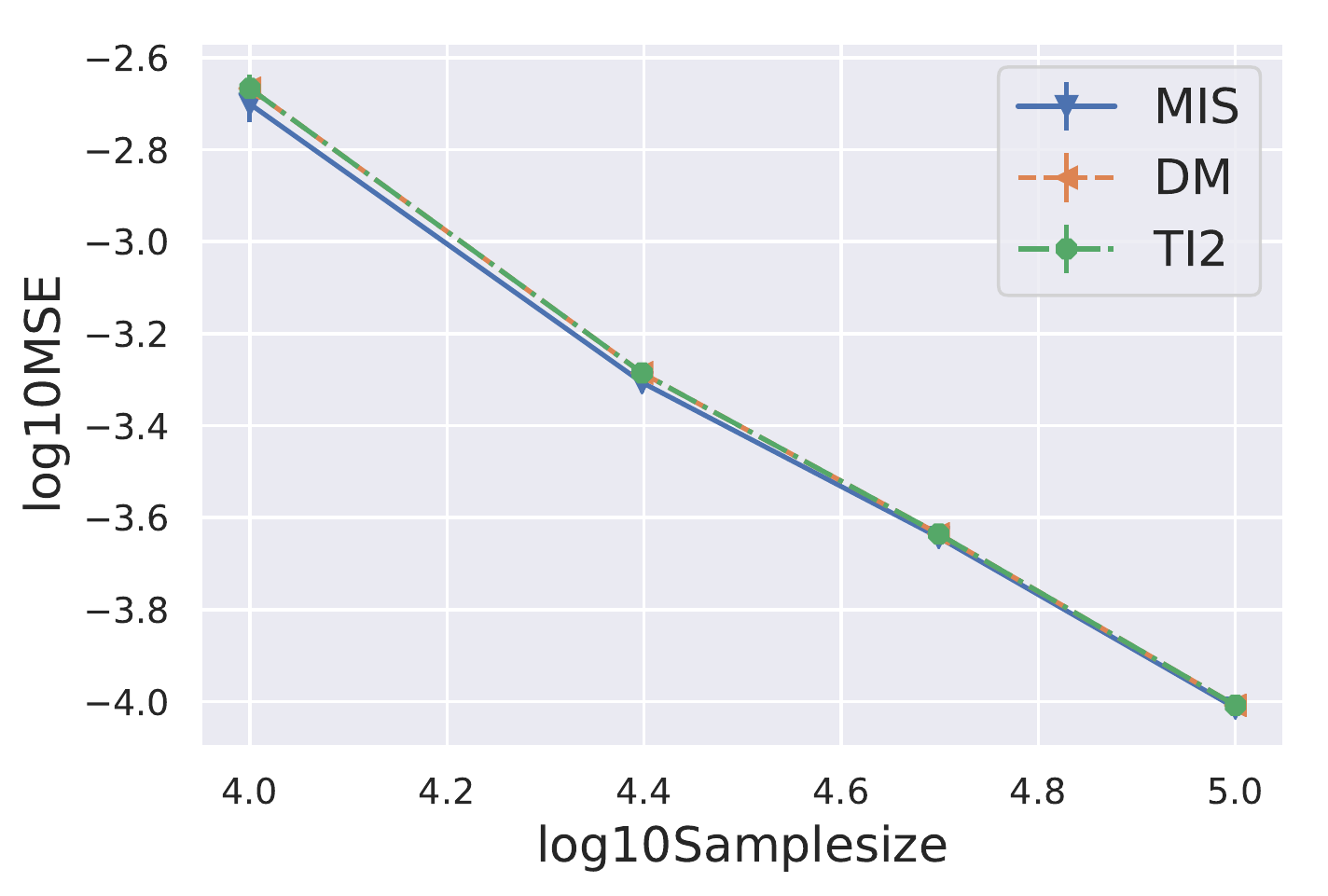}
\caption{Well-specified $q,w^{*}$}\label{fig:well1}
\end{minipage}
\begin{minipage}{0.325\textwidth}
\centering
\includegraphics[width=.95\linewidth]{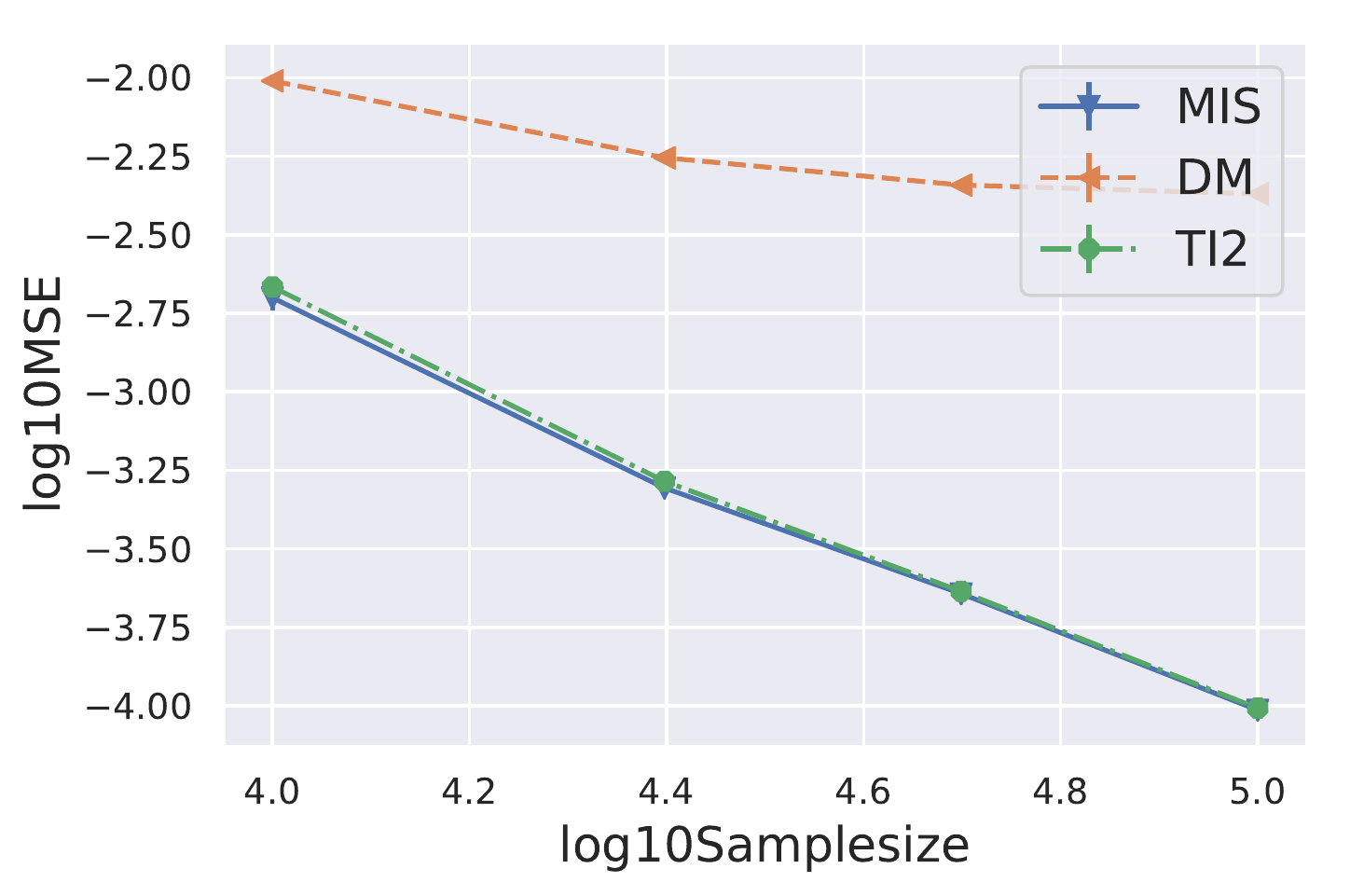}
\caption{Misspecified $q$}\label{Fig:misq1}
\end{minipage}%
\begin{minipage}{0.325\textwidth}
\centering
\includegraphics[width=.95\linewidth]{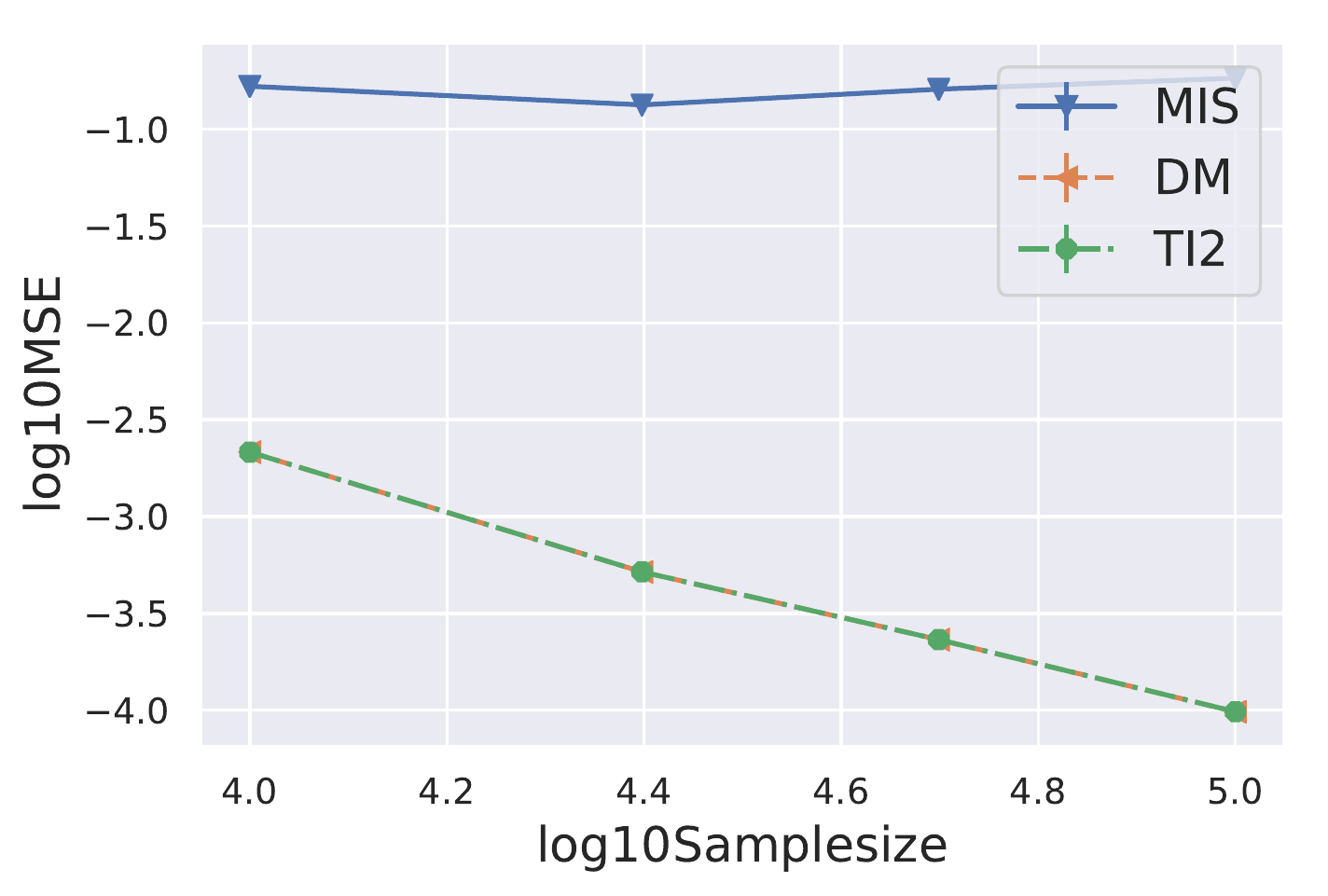}
\caption{Misspecified $w^{*}$}\label{Fig:misw1}
\end{minipage}\\%
\begin{minipage}{0.025\textwidth}
\rotatebox{90}{Modified Treatment}
\end{minipage}%
\begin{minipage}{0.33\textwidth}
\centering
\includegraphics[width=.95\linewidth]{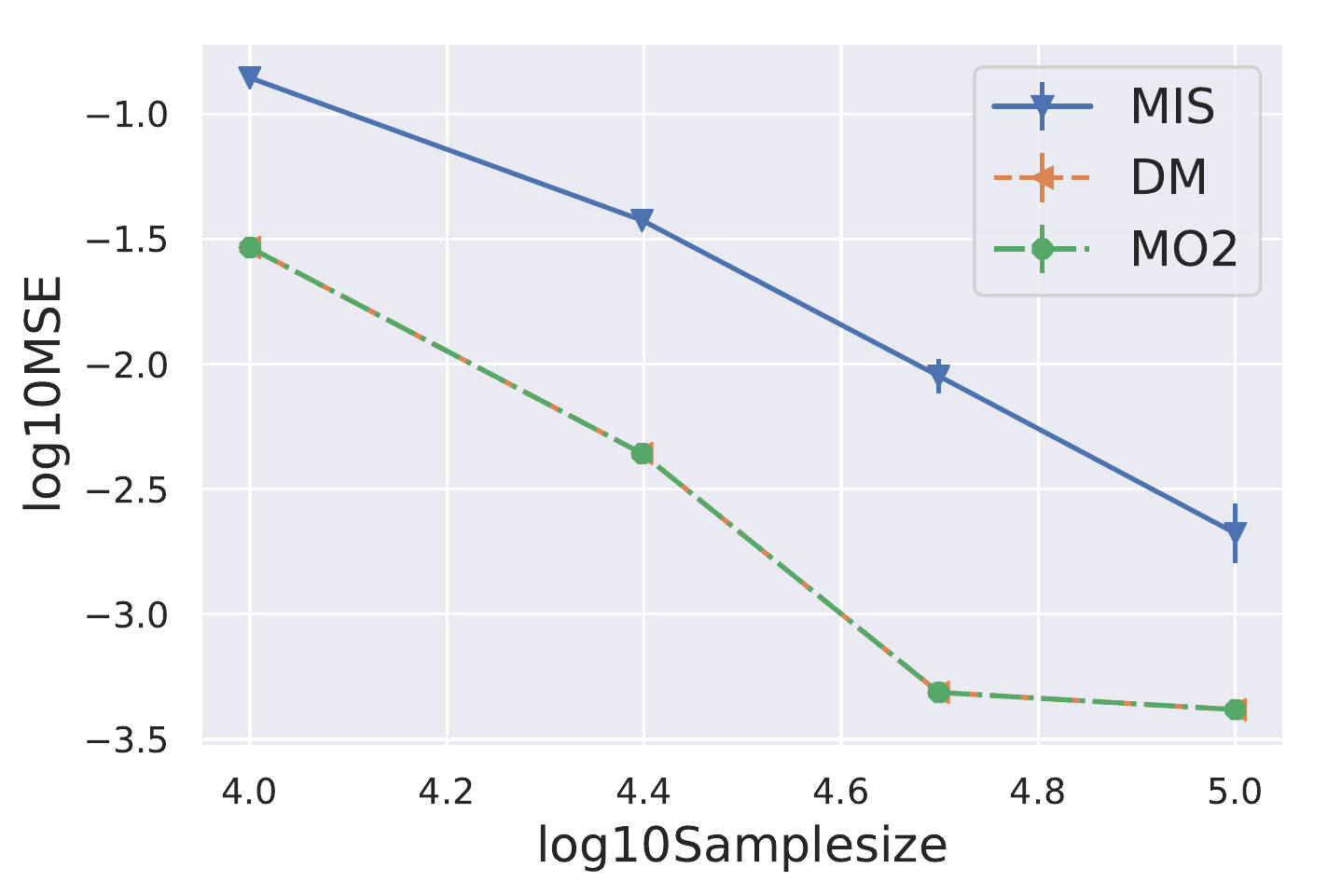}
\caption{Well-specified $q,w^{*}$}\label{fig:well2}
\end{minipage}
\begin{minipage}{0.33\textwidth}
\centering
\includegraphics[width=.95\linewidth]{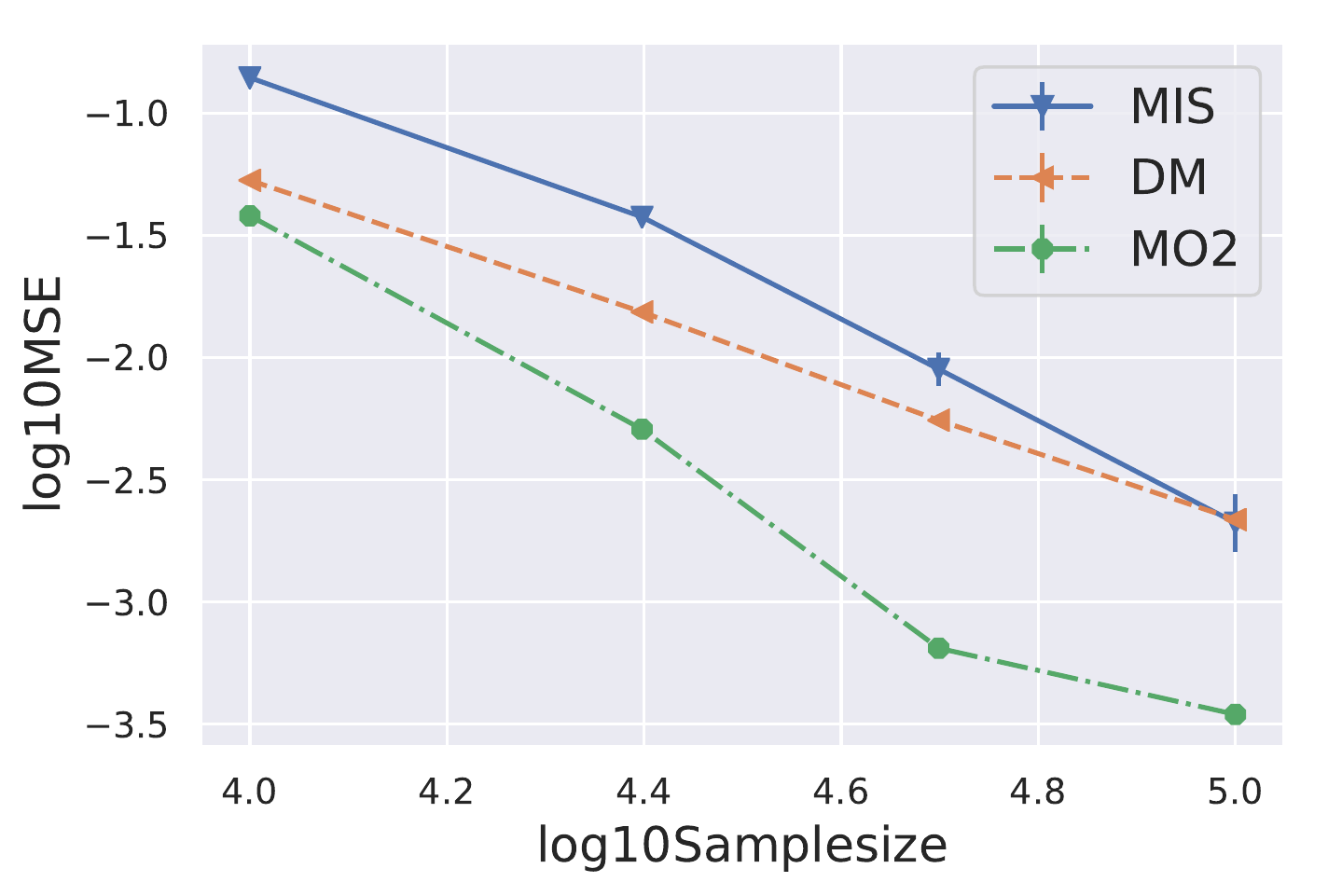}
\caption{Misspecified $q$}\label{Fig:misq2}
\end{minipage}%
\begin{minipage}{0.33\textwidth}
\centering
\includegraphics[width=.95\linewidth]{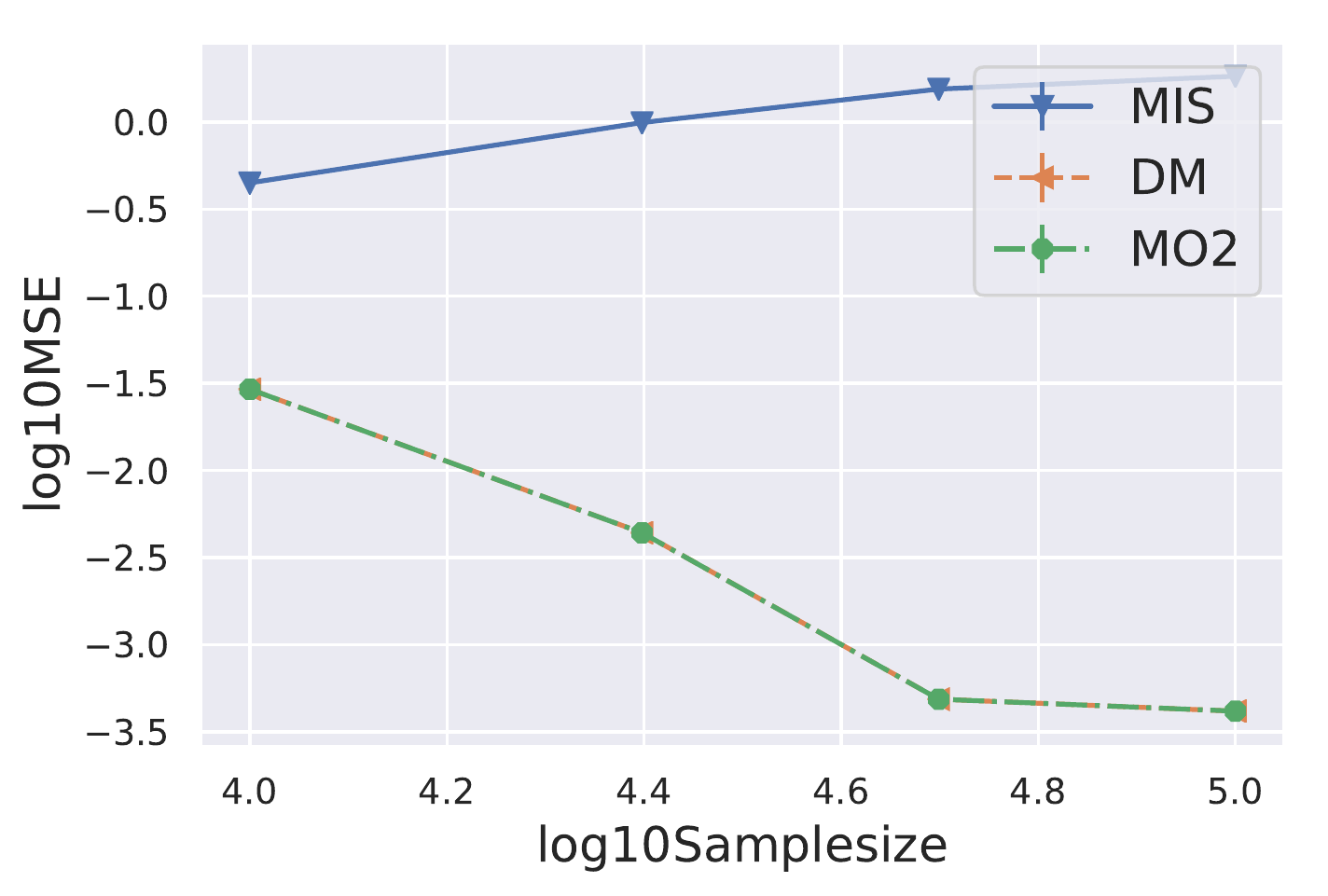}
\caption{Misspecified $w^{*}$}\label{fig:misw2}
\end{minipage}

\end{figure}

In this section we examine the performance of different OPE estimators in a time-invariant infinite-horizon setting. 
We use the Taxi environment, which is a commonly used tabular environment for OPE, which has $\Scal=\{1,\dots,2000\},\Acal=\{1,\dots,6\}$ (\citealp{DietterichT.G.2000HRLw}; we also refer the reader to \citealp[Section 5]{Liu2018} for more details), and we consider separate experiments for the case of tilting and modified treatment policies.
We consider observing of a single ($N=1$) trajectory of varying length $H$ ($\{1,2.5,5,10\}\times 10^4$). For each $H$ we run $60$ replications of the experiment.
We compare the stationary marginal IS estimator $\hat J_{\mathrm{MIS}}$ \citep{Liu2018}, the direct method $\hat J_{\mathrm{DM}}$, and one of our proposed estimators $\hat J_{\mathrm{TI2}},\hat J_{\mathrm{MO2}}$, depending on whether we are considering tilting or modified treatment policies.
We do not compare to step-wise IS \citep{precup2000eligibility} and DR \citep{jiang2016} as these estimators do not converge when given single-trajectory data \citep[as shown in][Section 7]{KallusNathan2019EBtC}.
Behavior and evaluation policies are set as follows. We run 150 iterations of $q$-learning to learn a near-optimal policy for the MDP and define this to be $\pi^b$.
We consider evaluating either a tilting policy with $u(a)=\lceil a/2\rceil$ or a modified treatment policy with $\tau(s,a)=(s+a)\,\mathrm{mod}\,6$. We set $\gamma=0.98$. We estimate $\bpol$ as 
$\hat\pi^b(a\mid s)=\sum_{i=1}^n\mathbb I\bracks{a^{(i)}=a,s^{(i)}=s}/\sum_{i=1}^n\mathbb I\bracks{s^{(i)}=s}$ 
and $w^{*}$- and $q$-functions by solving \cref{eq:nuisance1,eq:nuisance1b,eq:nuisance2} using $\{\mathbb I[s=i]:i=1,\dots,2000\}$ and $\{\mathbb I[s=i,a=j]:i=1,\dots,2000,j=1,\dots,6\}$ as test functions, respectively. We use these nuisance estimates to construct all estimators.
To validate double robustness, we also add Gaussian noise $\mathcal{N}(3.0,1.0)$ to either the $q$- or $w^{*}$-function estimates to simulate misspecification.
In \cref{fig:well1,Fig:misq1,Fig:misw1,fig:well2,Fig:misq2,fig:misw2}, we report the MSE of each estimator over the 60 replications with 95\% confidence intervals . 

We find the performance of $\hat J_{\mathrm{TI2}}$ or of $\hat J_{\mathrm{MO2}}$ is consistently good, with or without of model specification due to double robustness. 
While MIS and DM fail when their respective model is misspecified,
they do well when well-specified. 
Since either parametric misspecification or nonparametric rates for $w^{*}$ and $q$ is unavoidable in practice for large or continuous state-action spaces, $\hat J_{\mathrm{TI2}}$ and $\hat J_{\mathrm{MO2}}$ are seen to be superior to $\hat J_{\mathrm{DM}}$ and $\hat J_{\mathrm{MIS}}$. 

\section{Conclusions}\label{sec:conc}

We considered the evaluation of natural stochastic policies in RL, both in finite and infinite horizons. We derived the efficiency bounds and proposed estimators that achieved them under lax conditions on nuisance estimators that permit flexible machine learning methods. An important next question is \emph{learning} natural stochastic policies. This can perhaps be done using an off-policy policy gradient approach by extending \citet{EEOPG2020} to natural stochastic policies, where we take gradients in $u$ or $\tau$ that specify how and where we deviate from $\bpol$.


\bibliographystyle{chicago}
\bibliography{rc}

\newpage 

\bigskip
\begin{center}
{\large\bf SUPPLEMENTARY MATERIAL}
\end{center}

\appendix

\section{Proofs}

Throughout this paper, in an MDP, we define 
\begin{align*} 
    c(s)=1/\int u(a)\bpol(a|s)\rd a. 
\end{align*}
Then, $\epol(a|s)=u(a)\bpol(a|s)c(s)$. Similarly, in a TMDP, we define 
\begin{align*} 
    c_t(s_t)=1/\int u(a_t)\bpol(a_t|s_t)\rd a_t. 
\end{align*}
Then, $\pi^e_t(a_t|s_t)=u_t(a_t)\bpol_t(a_t|s_t)c_t(s_t)$. In addition, note that $\hat \pi^e(a|s)$ and $\hat \pi^e_t(a_t|s_t)$ are expressed as 
\begin{align*}
    \hat \pi^e(a|s) = u(a)\hat \pi^b(a|s)\hat c(s), \, \hat \pi^e_t(a_t|s_t) =u(a_t) \hat \pi^b(a_t|s_t)\hat c(s_t).  
\end{align*}

\subsection{Warm-up}
As a warm-up, we prove the results for the bandit case and NMDP case. In a bandit setting, we drop an index $t$. 

\begin{proof}[Proof of \cref{rem:bandit_example1}]
The entire regular parametric sumbodel is 
\begin{align*}
    \{p_{\theta}(s)p_{\theta}(a\mid s)p_{\theta}(r|s,a)\}, 
\end{align*}
where it matches with a true pdf at $\theta=0$. The score functions of the nonparametric model is decomposed as 
\begin{align*}
    g(\cj)&= \log p_{\theta}(s)+\log p_{\theta}(a\mid s)+ \log p_{\theta}(r|s,a)
    =g_{S}+g_{A|S}+g_{R|S,A}. 
\end{align*}
We calculate the gradient of the target functional $J(\epol)$ w.r.t. the nonparametric model. Since
\begin{align*}
    J(\epol)=\int rp_{\theta}(r\mid a,s)\epol(a\mid s;\bpol)p_{\theta}(s)\mathrm{d}(a,s,r), 
\end{align*}
we have 
\begin{align*}
   \nabla J(\epol)= &\E\left[\frac{\epol(A\mid S)}{\bpol(A\mid S)}\{R-q(S,A)\}g(\cj)+v(S)g(\cj)+R\frac{\nabla_{\theta}\epol(A\mid S;\bpol)}{\bpol(A\mid S)}\right].
\end{align*}
Especially, the third term is 
\begin{align*}
    &\E\left[ \frac{R\nabla_{\theta}\epol(A\mid S;\bpol)}{\bpol(A\mid S)}\right]\\
    &=\E\left[\left\{\E[R|S,A]\frac{u(A)}{c(S)}-\E\left[R \frac{u(A)}{c(S)}\mid S\right]\frac{u(A)}{c(S)}\right\}g_{A\mid S}\right] \\ 
    &=\E\left[\left\{\E[R\mid S,A]\frac{u(A)}{c(S)}-\E\left[R \frac{u(A)}{c(S)}\mid S\right]\frac{u(A)}{c(S)}\right\}g(\cj)\right] \\
    &=\E\left[\frac{\epol(A\mid S)}{\bpol(A\mid S)}\{q(S,A)-v(S)\}g(\cj)\right]. 
\end{align*}
Then, we have 
\begin{align*}
       \nabla J(\epol)=\E\left[\left\{\frac{\epol(A\mid S)}{\bpol(A\mid S)}(R-v(S))+v(S)\right\}g(\cj)\right]. 
\end{align*}
Since the gradient is unique for the current case, this concludes that the following is the EIF: 
\begin{align*}
    \frac{\epol(a|s)}{\bpol(a|s)}\{r-v(s)\}+v(s). 
\end{align*}
\end{proof}

\begin{proof}[Proof of \cref{rem:bandit_example2}]
The proof is also mentioned in \citet{MunozIvanDiaz2012PICE}. We add the proof here since this would improve the reader's understanding in the RL case. The entire regular parametric sumbodel is 
\begin{align*}
    \{p_{\theta}(s)p_{\theta}(a\mid s)p_{\theta}(r|s,a)\}
\end{align*}
where it matches with the true pdf at $\theta=0$. The score functions of the nonparametric model is decomposed as 
\begin{align*}
    g(\cj)&= \log p_{\theta}(s)+\log p_{\theta}(a\mid s)+ \log p_{\theta}(r|s,a)\\
    &=g_{S}+g_{A|S}+g_{R|S,A}. 
\end{align*}
We calculate the gradient of the target functional $J(\epol)$ w.r.t. the nonparametric model. Since
\begin{align*}
    J(\epol)&=\int rp_{\theta}(r\mid \tau(s,a),s) \bpol(a\mid s)p_{\theta}(s)\mathrm{d}(a,s,r)\\
    &=\int rp_{\theta}(r\mid a,s) \bpol(\tilde \tau(s,a)\mid s)\tilde \tau'(s,a)p_{\theta}(s)\mathrm{d}(a,s,r),  
\end{align*}
we have 
\begin{align*}
      \nabla_{\theta} J(\epol)=\E\left[\left\{\frac{\bpol(\tilde \tau(S,A)\mid S)\tilde \tau'(S,A)}{\bpol(A\mid S)}\{R-q(S,A)\}+\E[R\mid S,\tau(S,A)]\right\}g(\cj)\right]. 
\end{align*}
Since the gradient is unique for the current case, this concludes that the following is the EIF:
\begin{align*}
    \frac{\epol(a\mid s)}{\bpol(a\mid s)}\{r-q(s,a)\}+q(s,\tau(s,a)). 
\end{align*}
\end{proof}

\subsection{Proofs for \cref{sec:bounds}}

\begin{proof}[Proof of \cref{thm:mdp_example1}]

~
\paragraph*{Calculation of derivatives under a nonpametric TMDP}

The entire regular parametric submodel is 
\begin{align*}
    \{p_{\theta}(s_1)p_{\theta}(a_1|s_1)p_{\theta}(r_1|s_1,a_1)p_{\theta}(s_1|s_1,a_1)p_{\theta}(a_1|s_1)p_{\theta}(r_1|s_1,a_1)\cdots p_{\theta}(r_H|s_H,a_H)\}. 
\end{align*}
The score function of the parametric submodel is 
\begin{align*}
   g(\cjj_{r_H})&=\sum_{k=1}^{H}\nabla_{\theta} \log p_{\theta}(s_k\mid s_{k-1},a_{k-1})+\nabla_{\theta} \log p_{\theta}(a_{k}\mid s_{k})+\nabla_{\theta} \log p_{\theta}(r_k\mid s_{k},a_{k})\\
   &=\sum_{k=1}^{H}g_{S_{k}|S_{k-1},A_{k-1}}+\sum_{k=1}^{H}g_{A_{k}|S_k}+\sum_{k=1}^{H}g_{R_k|S_k,A_k}. 
\end{align*}

We have 
\begin{align*}
    &\nabla_{\theta} \mathrm{E}_{\pi^{e}}[\sum_{t=1}^{H}r_t]=\nabla_{\theta} \int \sum_{t=1}^{H}r_t \left\{\prod_{k=1}^{t}p_{\theta}(s_k|a_{k-1},s_{k-1})p_{\pi^{e}_k}(a_k|s_k;\theta)p_{\theta}(r_k|a_k,s_k) \right\}\mathrm{d}\mu(\cjj_{r_H})   \\
    &=\sum_{c=1}^{H}\{\E_{\pi^{e}}\left[(\E_{\pi^{e}}[r_c|s_1]-\E_{\pi^{e}}[r_c])g_{s_1}\right]+\E_{\pi^{e}}[(r_c-\E_{\pi^{e}}[r_c|s_c,a_c]) g_{R_c|S_c,A_c}] \\ 
    &+\E_{\pi^{e}}\left[\left(\E_{\pi^{e}}[\sum_{c=t+1}^{H} r_t|s_{c+1}]-\E_{\pi^{e}}[\sum_{c=t+1}^{H} r_t|s_{c},a_{c}]\right)g_{S_{c+1}|S_{c},A_{c}}\right]\\
    &+\E_{\bpol}\left[\mu_c(s_c,a_c)\left(\E_{\pi^{e}}[\sum_{c=t}^{H} r_t|s_{c},a_{c}]-\E_{\pi^{e}}[\sum_{c=t}^{H} r_t|s_{c}]\right)g_{A_c\mid S_c}\right] \}. 
\end{align*}
Except for the third term, the calculation is the same as Theorem 2 \citep{KallusUehara2019}. The third term has been calculated as the proof of \cref{thm:nmdp_example1}. Then, $\nabla_{\theta} \mathrm{E}_{\pi^{e}}[\sum_{t=1}^{H}r_t]$ is equal to 
\begin{align*}
    &\sum_{c=1}^{H}\{\E_{\bpol}[(\E[r_c|s_1]-\E_{\pi^{e}}[r_c])g]+\E_{\bpol}\left[\mu_c(s_c,a_c)(r_c-\E[r_c|s_c,a_c])g\right] \\
    & +\E_{\bpol}\left[\mu_c(s_c,a_c)(\E[\sum_{t=c+1}^{H} r_t|s_{c+1}]-\E[\sum_{t=c+1}^{H} r_t|s_{c},a_{c}])g\right] \\
     &+\E_{\bpol}\left[\mu_c(s_c,a_c)\left(\E_{\pi^{e}}[\sum_{c=t}^{H} r_t|s_{c},a_{c}]-\E_{\pi^{e}}[\sum_{c=t}^{H} r_t|s_{c}]\right)g\right]\}\\
    &= \E_{\bpol}\left[\left[-J+\sum_{c=1}^{H}\left \{\mu_c(s_c,a_c)\left\{r_c-\E_{\pi_e}[\sum_{t=c}^{H}r_t|s_c]\right\}+\mu_{c-1}(s_{c-1},a_{c-1})\sum_{t=c}^{H}\E_{\pi_e}[r_t|s_c]\right \}\right]g(\mathcal{J}_{r_H})\right]
\end{align*}
In the end, we can conclude that the following is a derivative:
\begin{align}
\label{eq:gradient_example1_mdp}
    -J+\sum_{c=1}^H\mu_c\{r_c-v_{c}(s_c)\}+\mu_{c-1}v_{c}(s_c).
\end{align}

\paragraph*{Projection onto the tangent space}

Then, based on the argument in Appendix B \citep{EEOPG2020}, we need to project it onto the tangent space spanned by the nonparametric model deduced by an MDP. Writing the gradient \cref{eq:gradient_example1_mdp} as $\phi$, the projection of $\phi$ onto the tangent space is calculated as follows:
\begin{align*}
    & \sum_{j=1}^H \E[\phi |R_j,S_j,A_j]-\E[\phi |S_j,A_j]+ \E[\phi |A_j,S_j]-\E[\phi |S_j]+\E[\phi |S_j,A_{j-1},S_{j-1}]-\E[\phi |A_{j-1},S_{j-1}]\\
    &=\{\sum_{j=1}^H  \mu_j (R_j-\E[R_j\mid S_j,A_j])\}\\
    &+\E[\sum_{c=j}^H\mu_c\{R_c-v_c(S_c)+v_{c+1}(S_{c+1})\}\mid A_j,S_j]-\E[\sum_{c=j}^H\mu_c\{R_c-v_c(S_c)+v_{c+1}(S_{c+1})\}\mid S_j]\\
  &+\E[\sum_{c=j}^H\mu_c\{R_c-v_c(S_c)+v_{c+1}(S_{c+1})\}\mid S_j] -\E[\sum_{c=j}^H\mu_c\{R_c-v_c(S_c)
  +v_{c+1}(S_{c+1})\mid S_{j-1},A_{j-1}]\\
  &+\mu_{j-1}v_{j}(S_{j})-\E[\mu_{j-1} v_{j}(S_{j})\mid S_{j-1},A_{j-1}]\\
      &=\{\sum_{j=1}^H  \mu_j (R_j-\E[R_j\mid S_j,A_j])\}+\mu_j\{\E[R_j\mid S_j,A_j]-v_j(S_j)+\E[v_{j+1}(S_{j+1})|S_j,A_j]\} \\
  &+\mu_{j-1}v_{j}(S_{j})-\E[\mu_{j-1} v_{j}(S_{j})\mid S_{j-1},A_{j-1}]\\
  &=-J+v_1(S_1)+\sum_{j=1}^H  \mu_j(S_j,A_j)\{R_j-v_{j}(S_j)+v_{j+1}(S_{j+1})\}=\phi.  
\end{align*}
This concludes that $\phi$ is actually the EIF. 

\paragraph*{Efficiency bound}

From the law of the total variance, the efficiency bound is 
\begin{align*}
    &\var\left[ \sum_{c=1}^H\mu_c\{R_c-v_{c}(S_c)\}+\mu_{c-1}v_{c}(S_c) \right]\\
    &=\sum_{t=0}^H  \E[\var[\E[  \sum_{c=1}^H\mu_c\{R_c+v_{c+1}(S_{c+1})-v_{c}(S_c)\} |\cj_{S_{t+1}}]|\cj_{S_{t}}] ] \\
    &=\sum_{t=0}^{H}\E[\var[\mu_t\{R_t+v_{t+1}(S_{t+1})\}\mid \cj_{S_t}]]
    = \sum_{t=0}^{H}\E[w^2_t(S_t)\var[\eta_t(S_t,A_t)\{R_t+v_{t+1}(S_{t+1})\}\mid S_t]]. 
\end{align*}

\paragraph*{Order of the efficiency bounds}

We use importance sampling as follows, 
\begin{align*}
    &\sum_{t=0}^{H}\E_{\bpol}[w^2_t\var_{\bpol}[\lambda_t\{r_t+v_t(s_{t+1})\}\mid s_t]]=\sum_{t=0}^{H}\E_{\bpol}[w^2_t \E_{\bpol}[\lambda^2_t\{r_t+v_t(s_{t+1})-v_t(s_t)\}^2  \mid s_t]]\\
    &\leq CC' \sum_{t=0}^{H}\E_{\epol}[ \E_{\epol}[\{r_t+v_t(s_{t+1})-v_t(s_t)\}^2  \mid s_t]]
    = CC' \sum_{t=0}^{H}\E_{\epol}[ \{r_t+v_t(s_{t+1})-v_t(s_t)\}^2 ]\\
      &= CC' \E_{\epol}[ \{\sum_{t=0}^{H}r_t+v_t(s_{t+1})-v_t(s_t)\}^2 ]  = CC' \E_{\epol}[ \{[\sum_{t=1}^{H}r_t]-J\}^2]\leq CC'(R_{\max}H)^2. 
\end{align*}

\end{proof}
 
\begin{remark}
Noting 
\begin{align*}
    \E[\var[f(Z)|X]]=\var[\E[f(Z)|X,Y]|X]+\E[\var[f(Z)|X,Y]|X]. 
\end{align*}
for random variables $X,Y,Z$, the difference of this efficiency bound and the efficiency bound of the pre-specified evaluation policies is 
\begin{align*}
    \sum_{t=0}^H\E[w^2_t(S_t)\var[\eta_t(S_t,A_t)q_t(S_t,A_t)\mid S_t]]. 
\end{align*}
\end{remark}

\begin{proof}[Proof of \cref{thm:mdp_example2}]

~
\paragraph*{EIF and efficiency bound under a TMDP}

The entire regular parametric submodel is 
\begin{align*}
    \{p_{\theta}(s_1)p_{\theta}(a_1|s_1)p_{\theta}(r_1|s_1,a_1)p_{\theta}(s_1|s_1,a_1)p_{\theta}(a_1|s_1)p_{\theta}(r_1|s_1,a_1)\cdots p_{\theta}(r_H|s_H,a_H)\},
\end{align*}
where it matches with the true pdf at $\theta=0$. The score function of the parametric submodel is 
\begin{align*}
   g(j_{r_H})&=\sum_{k=1}^{H}\nabla_{\theta} \log p_{\theta}(s_k\mid s_{k-1},a_{k-1})+\nabla_{\theta} \log p_{\theta}(a_{k}\mid s_{k-1})+\nabla_{\theta} \log p_{\theta}(r_k\mid s_{k},a_{k})\\
   &=\sum_{k=1}^{H}g_{S_{k}|S_{k-1},A_{k-1}}+\sum_{k=1}^{H}g_{A_{k}|S_{k-1}}+\sum_{k=1}^{H}g_{R_k|S_k,A_k}. 
\end{align*}
The target functional is 
\begin{align*}
&\int \sum_{t=1}^{H}r_t \left\{\prod_{k=1}^{t}p_{\theta}(s_k|\tau(a_{k-1},s_{k-1}),s_{k-1})\bpol_k(a_k|s_k;\theta)p_{\theta}(r_k|\tau(a_k,s_k),s_k) \right\}\mathrm{d}\mu(\cjj_{r_H})\\
&=\int \sum_{t=1}^{H}r_t \left\{\prod_{k=1}^{t}p_{\theta}(s_k|a_{k-1},s_{k-1})\pi^{e}_k(a_k|s_k;\theta)p_{\theta}(r_k|a_k,s_k) \right\}\mathrm{d}\mu(\cjj_{r_H}),
\end{align*}
where $\epol_k(a_k \mid s_k)$ is $\bpol_k(\tilde \tau(s,a)\mid s)\tilde \tau'(s,a)$, $\tilde \tau_k(\cdot,s)$ is the inverse function of $\tau_k(\cdot,s)$. 
Here, we use a change of variables: $\tau(a_{k-1},s_{k-1})=u_{k-1}$, and write $u_k$ as $a_k$. We have 
\begin{align*}
    &\nabla_{\theta} \mathrm{E}_{\pi^{e}}[\sum_{t=1}^{H}r_t] =\nabla_{\theta} \int \sum_{t=1}^{H}r_t \left\{\prod_{k=1}^{t}p_{\theta}(s_k|\tau(s_{k-1},a_{k-1}),s_{k-1})\pi^{e}_k(a_k|s_k;\theta)p_{\theta}(r_k|a_k,s_k) \right\}\mathrm{d}\mu(\cjj_{r_H})   \\
    &=\sum_{c=1}^{H}\{\E_{\pi^{e}}\left[(\E_{\pi^{e}}[r_c|s_1]-\E_{\pi^{e}}[r_c])g_{S_1}\right]+\E_{\pi^{e}}[(r_c-\E_{\pi^{e}}[r_c|s_c,a_c]) g_{R_c|S_c,A_c}] \\ 
    &+\E_{\pi^{e}}\left[\left(\E_{\pi^{e}}[\sum_{c=t+1}^{H} r_t|s_{c+1}]-\E_{\pi^{e}}[\sum_{c=t+1}^{H} r_t|s_{c},a_{c}]\right)g_{S_{c+1}|S_{c},A_{c}}\right]\\
    &+\E_{\bpol}\left[\prod_{k=1}^{t-1}\eta_k \left(\E_{\pi^{e}}[\sum_{c=t}^{H} r_t|s_{c},\tau(s_c,a_{c})]-\E_{\pi^{e}}[\sum_{c=t}^{H} r_t|s_{c}]\right)g_{A_c\mid S_c}\right]. 
\end{align*}
Except for the third line, the proof is almost the same as that of \citet[Theorem 2]{KallusUehara2019}. The third line is proved by
{\small 
\begin{align*}
    &\sum_{c=1}^{H} \int \sum_{t=1}^{H}r_t \left\{\prod_{k\neq c}^{t}p(s_k|a_{k-1},s_{k-1})p_{\pi^{e}_k}(a_k|s_k)p(r_k|a_k,s_k) \right\} p(s_c|\tau(a_{c-1},s_{c-1}),s_{c-1}) \\ & \,\,\,\,\, \times \nabla \bpol(a_c|s_c;\theta)p(r_c|\tau(a_c,s_c),s_c) \mathrm{d}\mu(\cjj_{r_H})\\
   &=   \sum_{c=1}^{H} \int \sum_{t=c}^{H}r_t \left\{\prod_{k\neq c}^{t}p(s_k|a_{k-1},s_{k-1})p_{\pi^{e}_k}(a_k|s_k)p(r_k|a_k,s_k) \right\} p(s_c|\tau(a_{c-1},s_{c-1}),s_{c-1}) \\ & \,\,\,\,\, \times \bpol(a_c|s_c)p(r_c|\tau(a_c,s_c),s_c)g_{A_c|S_c} \mathrm{d}\mu(\cjj_{r_H}) \\
   &=\E_{\bpol}\left[\prod_{k=1}^{t-1}\eta_k \E_{\pi^{e}}[\sum_{c=t}^{H} r_t|s_{c},\tau(s_c,a_{c})]g_{A_c\mid S_c}\right]\\
   &=\E_{\bpol}\left[\prod_{k=1}^{t-1}\eta_k \left(\E_{\pi^{e}}[\sum_{c=t}^{H} r_t|s_{c},\tau(s_c,a_{c})]-\E_{\pi^{e}}[\sum_{c=t}^{H} r_t|s_{c}]\right)g_{A_c\mid S_c}\right].
\end{align*}
}

Then, 
\begin{align*}
    &\nabla_{\theta} \mathrm{E}_{\pi^{e}}[\sum_{t=1}^{H}r_t] =  \sum_{c=1}^{H}\{\E_{\bpol}[(\E_{\epol}[r_c|s_1]-\E_{\pi^{e}}[r_c])g]+\E_{\bpol}\left[\mu_c(s_c,a_c)(r_c-\E[r_c|s_c,a_c])g\right] \\
    & +\E_{\bpol}\left[\mu_c([\E_{\epol}[\sum_{t=c+1}^{H} r_t|s_{c+1}]-\E_{\epol}[\sum_{t=c+1}^{H} r_t|s_{c},a_{c}])g\right]\\
     &+\E_{\bpol}\left[\mu_{c-1} \left(\E_{\pi^{e}}[\sum_{c=t}^{H} r_t|s_{c},\tau(s_c,a_{c})]-\E_{\pi^{e}}[\sum_{c=t}^{H} r_t|s_{c}]\right)g\right]\}.
\end{align*}     
Here, note that we define $\E_{\bpol}[\prod_{k=1}^t \eta_k \mid s_t,a_t]=\mu_t$. Then, $\nabla_{\theta} \mathrm{E}_{\pi^{e}}[\sum_{t=1}^{H}r_t]$ is equal to 
\begin{align*}
    &= \E \left[\left[-J+\sum_{c=1}^{H}\left \{\mu_c \left\{r_c-\E_{\pi_e}[\sum_{t=c}^{H}r_t|s_c,a_c]\right\}+\mu_{c-1}\sum_{t=c}^{H}\E_{\pi_e}[r_t|s_c]\right \}\right]g(\mathcal{J}_{r_H})\right]\\
    &+\E\left[\mu_{c-1}\left(\E_{\pi^{e}}[\sum_{c=t}^{H} r_t|s_{c},\tau(s_c,a_{c})]-\E_{\pi^{e}}[\sum_{c=t}^{H} r_t|s_{c}]\right)g\right] \\ 
    &= \E \left[\left[-J+\sum_{c=1}^{H}\left \{\mu_c\left\{r_c-\E_{\pi_e}[\sum_{t=c}^{H}r_t|s_c,s_c]\right\}+\mu_{c-1}\sum_{t=c}^{H}\E_{\pi_e}[r_t|s_c,\tau(a_c,s_c)]\right \}\right]g(\mathcal{J}_{r_H})\right]. 
\end{align*}
Then, as in the proof of \cref{thm:mdp_example1} via the projection onto the tangent space, the EIF becomes 
\begin{align*}
    -J(\epol)+\sum_{c=1}^H\mu_c\{r_c-q_c(s_c,a_c)\}+\mu_{c-1}q_c(s_c,\tau(a_c,s_c)). 
\end{align*}
In addition, as in the proof of \cref{thm:mdp_example1}, the efficiency bound is 
\begin{align*}
    \sum_{c=0}^H \E[\mu^2_c(S_c,A_c) \var[R_c+q_{c+1}(S_c,\tau(A_c,S_c))\mid S_c,A_c]]. 
\end{align*}

\paragraph*{Order of the efficiency bound}

First, we observe
\begin{align}
\label{eq:observation}
    \var[R_c+q_{c+1}(S_{c+1},\tau(A_{c+1},S_{c+1}))\mid S_c=s_c,A_c=a_c]=\var_{\epol}[r_c+q_{c+1}(s_{c+1},a_{c+1})\mid s_c,a_c]. 
\end{align}
This is proved by 
{\small 
\begin{align*}
    &\var[R_c+q_{c+1}(S_{c+1},\tau(A_{c+1},S_{c+1}))\mid S_c=s_c,A_c=a_c]\\
    &=\int \{r_c+q_{c+1}(s_{c+1},\tau(a_{c+1},s_{c+1}))-q_c(s_c,a_c)\}^2 p(r_c\mid s_c,a_c)p(s_{c+1}\mid s_c,a_c)\bpol(a_{c+1}\mid s_{c+1})\mathrm{d}(r_c,s_{c+1},a_{c+1})\\
    &=\int \{r_c+q_{c+1}(s_{c+1},u_{c+1}))-q_c(s_c,a_c)\}^2 p(r_c\mid s_c,a_c)p(s_{c+1}\mid s_c,a_c)\times \\
    &\,\,\,\bpol(\tilde \tau(u_{c+1},s_{c+1})\mid s_{c+1})\tilde \tau'(u_{c+1},s_{c+1})\mathrm{d}(r_c,s_{c+1},u_{c+1})\\
    &=\int \{r_c+q_{c+1}(s_{c+1},u_{c+1}))-q_c(s_c,a_c)\}^2 p(r_c\mid s_c,a_c)p(s_{c+1}\mid s_c,a_c)\epol(u_{c+1} \mid s_{c+1})\mathrm{d}(r_c,s_{c+1},u_{c+1})\\
    &=\var_{\epol}[r_c+q_{c+1}(s_{c+1},a_{c+1}))\mid s_c,a_c]. 
\end{align*}
}

Then, we have
\begin{align*}
    &\sum_{c=0}^H \E[\mu^2_c(S_c,A_c) \var[R_c+q_{c+1}(S_c,\tau(A_{c+1},S_{c+1}))\mid S_c,A_c]]\\
    &\leq CC'\sum_{c=0}^H \E_{\epol}[ \var_{\bpol}[r_c+q_{c+1}(s_{c+1},\tau(a_{c+1},s_{c+1}))\mid s_c,a_c]]\\
  &= CC'\sum_{c=0}^H \E_{\epol}[ \var_{\epol}[r_c+q_{c+1}(a_{c+1},a_{c+1}))\mid s_c,a_c]]\\
&= CC'\sum_{c=0}^H \E_{\epol}[\{r_c+q_{c+1}(s_{c+1},a_{c+1})-q_{c}(s_c,a_c)\}^2]\\
&=  CC' \E_{\epol}[\{\sum_{c=0}^H r_c+q_{c+1}(s_{c+1},a_{c+1})-q_{c}(s_c,a_c)\}^2]= C C' \E_{\epol}[\{[\sum_{c=1}^H r_c]-J\}^2]\\
&\leq CC'R^2_{\max}H^2. 
\end{align*}
From the first line to the second line, we use $\mu_c\leq CC'$. From the second line to the third line, we use \cref{eq:observation}. From the fourth line to the fifth line, we use a fact that the cross-term is equal to $0$. 
\end{proof}

\begin{remark}
Noting 
\begin{align*}
    \E[\var[f(Z)|X]]=\var[\E[f(Z)|X,Y]|X]+\E[\var[f(Z)|X,Y]|X], 
\end{align*}
for random variables $X,Y,Z$, the difference regarding the efficiency bound between the above and that of the pre-specified evaluation policy is 
\begin{align*}
 &  \sum_{t=1}^H\E[\mu^2_t(S_t,A_t)\var[R_t+q_{t+1}(S_{t+1},\tau(S_{t+1},A_{t+1}) )\mid S_{t+1},R_{t},S_t,A_t]] \\ 
   &=\sum_{t=1}^H\E[\mu^2_t(S_t,A_t)\var[q_{t+1}(S_{t+1},\tau(S_{t+1},A_{t+1})) \mid S_{t+1}]]. 
\end{align*}
\end{remark}

\begin{proof}[Proof of \cref{thm:efficiency_tmdp1}]
For simplicitly, we consider the case $K=2$. First, we prove
\begin{align}
   & \P_{\cu_1}[\phi(\hat w^{(1)},\hat \pi^{b(1)},\hat q^{(1)} ) |\Lcal_1]-\P_{\cu_1}[\phi(w, \pi^{b}, q ) |\Lcal_1] \nonumber \\
    &= \bG_{\cu_1}[ \phi(\hat w^{(1)},\hat \pi^{b(1)},\hat q^{(1)} )- \phi(w, \pi^{b}, q )  ]  \label{eq:f1_tmdp1}\\
    &+\E[\phi(\hat w^{(1)},\hat \pi^{b(1)},\hat q^{(1)} ) |\Lcal_1]-\E[\phi(w, \pi^{b}, q ) |\Lcal_1]   \label{eq:f2_tmdp1}\\
    &= \op(n^{-1/2}). \nonumber
\end{align}
From now on, we drop an index (1) for simplicity. In addition, we use a notation of $c_t(s_t),\hat c_t(s_t)$. 

~
\paragraph{Part1: \cref{eq:f2_tmdp1} is $\op(1/\sqrt{n})$}
~
First, we consider controlling the following term:
$$
\E[\sum_{t=1}^H \hat w_t(S_t)\hat \eta_t(S_t,A_t)\{R_t-\hat q_t(S_t,A_t)\}+ \hat w_{t-1}(S_{t-1})\hat \eta_{t-1}(S_{t-1},A_{t-1})\hat v_t(S_t)|\Lcal_1]-J. 
$$
By some algebra, we have 
\begin{align}
    &\E[\sum_{t=1}^H \hat w_t(S_t)\hat \eta_t(S_t,A_t)\{R_t-\hat q_t(S_t,A_t)\}+ \hat w_{t-1}(S_{t-1})\hat \eta_{t-1}(S_{t-1},A_{t-1})\hat v_t(S_t)|\Lcal_1]-J \nonumber \\
    &= \E[\sum_{t=1}^H \hat w_t(S_t)\hat \eta_t(S_t,A_t)\{R_t-\hat q_t(S_t,A_t)\}  \nonumber \\
    &+\E_{\epol}[\hat q_t(s_t)|A_t]|\Lcal_1]-J+\E[\sum_{t=1}^H  \hat w_{t-1}(S_{t-1})\hat \eta_{t-1}(S_{t-1},A_{t-1})\{\hat v_t(S_t)-\E_{\epol}[\hat q_t(s_t,A_t)|A_t]\} |\Lcal_1]   \nonumber \\
    &=\E[\sum_{t=1}^H \{\hat w_t(S_t)\hat \eta_t(S_t,A_t)-w_t(S_t)\eta_t(S_t,A_t) \}\{\hat q_t(S_t,A_t)-q_t(S_t,A_t) \} |\Lcal_1 ]  \label{eq:part1_1}\\
    &+\E[\sum_{t=1}^H \{\hat w_{t-1}(S_{t-1})\hat \eta_{t-1}(S_{t-1},A_{t-1})-w_t(S_{t-1})\eta_{t-1}(S_{t-1},A_{t-1}) \}\{-\hat v_t(S_t)+v_t(S_t) \} |\Lcal_1 ]   \nonumber \\
    &+\E[\sum_{t=1}^H \mu_t(S_t,A_t)\{-\hat q_t(S_t,A_t)+q_t(S_t,A_t)\}+\mu_{t-1}(S_{t-1},A_{t-1})\{\hat v_{t}(S_{t})-v_{t}(S_t)\}|\Lcal_1]  \label{eq:part1_3} \\
    &+\E[\sum_{t=1}^H \{\hat \mu_t(S_t,A_t)-\mu_t(S_t,A_t)\}\{R_t-q_{t}(S_t,A_t)+v_{t+1}(S_{t+1})\} \}|\Lcal_1]  \label{eq:part1_2}\\
    &+\E[\sum_{t=1}^H  \hat w_{t-1}(S_{t-1})\hat \eta_{t-1}(S_{t-1},A_{t-1})\int \{\hat \pi^e_t(a_t|S_t)-\pi^e_t(a_t|S_t)\}\hat q_t(a_t,S_t)\mathrm{d}a_t|\Lcal_1].  \nonumber 
\end{align}   
Here, the terms from \cref{eq:part1_1} to \cref{eq:part1_2} are the same as the decomposition in the same as Theorem 2 \citep{KallusUehara2019}. Since  \cref{eq:part1_2} and  \cref{eq:part1_3} are equal to $0$, the above is equal to 
{\small 
 \begin{align*}  
    &\E[\sum_{t=1}^H \{\hat w_t(S_t)\hat \eta_t(S_t,A_t)-w_t(S_t)\eta_t(S_t,A_t) \}\{\hat q_t(S_t,A_t)-q_t(S_t,A_t) \}  |\Lcal_1]\\
    &+\E[\sum_{t=1}^H \{\hat w_{t-1}(S_{t-1})\hat \eta_{t-1}(S_{t-1},A_{t-1})-w_t(S_{t-1})\eta_{t-1}(S_{t-1},A_{t-1}) \}\{-\hat v_t(S_t)+v_t(S_t) \}  |\Lcal_1]\\
    &+\E[\sum_{t=1}^H  \{\hat w_{t-1}(S_{t-1})\hat \eta_{t-1}(S_{t-1},A_{t-1})-w_{t-1}(S_{t-1}) \eta_{t-1}(S_{t-1},A_{t-1})\}\int \{\hat \pi^e_t(a_t|S_t)-\pi^e_t(a_t|S_t)\}\hat q_t(a_t,S_t)\mathrm{d}a_t|\Lcal_1]\\
    &+\E[\sum_{t=1}^H  w_{t-1}(S_{t-1})\eta_{t-1}(S_{t-1},A_{t-1})\int \{\hat \pi^e_t(a_t|S_t)-\pi^e_t(a_t|S_t)\}\hat q_t(a_t,S_t)\mathrm{d}a_t|\Lcal_1]\\
    &=\alpha_1\beta+\alpha_2\beta+(\alpha_1+\alpha_2)\alpha_2 +\E[\sum_{t=1}^H  w_t(S_t)\int \{\hat \pi^e_t(a_t|S_t)-\pi^e_t(a_t|S_t)\}\hat q_t(a_t,S_t)\mathrm{d}a_t|\Lcal_1]. 
\end{align*}
}
Here, we use $\|\hat v_t-v_t\|_2\leq C\|\hat q_t-q_t\|_2\lesssim \beta$. Next, we also use a fact $\|\hat c_t(S_t)-c_t(S_t)\|^2_2=\alpha^2_2$ since 
\begin{align*}
    \|\hat c_t(S_t)-c_t(S_t)\|^2_2&=\int \left\{ u(a_t)\hat \pi^b_t(a_t|s_t)-\pi^b_t(a_t|s_t)u(a_t)\right\}^2 p_{\bpol}(s_t)\mathrm{d}\mu(a_t,s_t)\\
    &\lessapprox \| \hat \pi^b_t(A_t|S_t)-\pi^b_t(A_t|S_t)\|^2_2. 
\end{align*}
Therefore, noting what we want to control $$\E[\sum_{t=1}^H \hat w_t(S_t)\hat \eta_t(S_t,A_t)\{R_t-\hat v_t(S_t)\}+\hat w_{t-1}(S_{t-1})\hat \eta_{t-1}(S_{t-1},A_{t-1})\hat v_t(S_t)|\Lcal_1]-J,$$ the following holds: 
\begin{align*}
    &\E[\sum_{t=1}^H \hat w_t(S_t)\hat \eta_t(S_t,A_t)\{R_t-\hat v_t(S_t)\}+\hat w_{t-1}(S_{t-1})\hat \eta_{t-1}(S_{t-1},A_{t-1})\hat v_t(S_t)|\Lcal_1]-J\\
    &=\E\left[\sum_{t=1}^H \hat w_t(S_t)\hat \eta_t(S_t,A_t) \{-\hat v_t(S_t)+\hat q_t(S_t,A_t)\} + w_t(S_t)\int \{\hat \pi^e_t(a_t|S_t)-\pi^e_t(a_t|S_t)\}\hat q_t(a_t,S_t)\mathrm{d}a_t|\Lcal_1 \right]\\
    &+\alpha_1\beta+\alpha_2\beta+\alpha_1\alpha_2 +\alpha^2_2. 
\end{align*}

Then, the main term in the above is further expanded as follows.
{\small 
\begin{align*}
    &\E\left[\sum_{t=1}^H w_t(S_t)\hat \eta_t(S_t,A_t) \{-\hat v_t(S_t)+\hat q_t(S_t,A_t)\}|\Lcal_1 \right]+\E[\sum_{t=1}^H  w_t(S_t)\int \{\hat \pi^e_t(a_t|S_t)-\pi^e_t(a_t|S_t)\}\hat q_t(a_t,S_t)\mathrm{d}a_t|\Lcal_1]\\
    &=\E\left[\sum_{t=1}^H w_t(S_t)\hat c_t(S_t)u_t(A_t)\{-\hat c_t(S_t)\int u_t(a_t)\hat q_t(a_t,S_t)\hat \pi^b_t(a_t|S_t)\mathrm{d}a_t +\hat q_t(S_t,A_t)\}|\Lcal_1 \right]\\
    &+\E[\sum_{t=1}^H w_t(S_t)\int u_t(a_t)\{\hat c_t(S_t)\hat \pi^b_t(a_t|S_t) -c_t(S_t)\pi^b_t(a_t|S_t) \}\hat q_t(a_t,S_t)\mathrm{d}a_t|\Lcal_1]\\
  &=\E\left[\sum_{t=1}^H w_t(S_t) \int \hat c_t(S_t)u_t(g_t)\bpol_t(g_t|S_t)\{-\hat c_t(S_t)\int u_t(a_t)\hat q_t(a_t,S_t)\hat \pi^b_t(a_t|S_t)\mathrm{d}a_t +\hat q_t(g_t,S_t)\}\mathrm{d}g_t|\Lcal_1 \right]\\
  &+\E[\sum_{t=1}^H w_t(S_t)\int u_t(a_t)\{\hat c_t(S_t)\hat \pi^b_t(a_t|S_t) -c_t(S_t)\pi^b_t(a_t|S_t) \}\hat q_t(a_t,S_t)\mathrm{d}a_t]|\Lcal_1]. 
\end{align*}
}
Noting we have $1/c_t(S_t)=\int u_t(g_t)\bpol_t(g_t|S_t)dg_t$, the following holds: 
\begin{align*}
  &=\E[\sum_{t=1}^H w_t(S_t)\int \{\hat c_t(S_t)-c_t(S_t)\}u_t(g_t)\bpol_t(g_t|S_t)\hat q_t(g_t,S_t)\mathrm{d}g_t |\Lcal_1 ]+\\
  &+\E[\sum_{t=1}^H w_t(S_t)\{-\hat c^2_t(S_t)/c_t(S_t)+\hat c_t(S_t) \}\int u_t(a_t)\hat \pi^b_t(a_t|S_t)\hat q_t(a_t,S_t)\mathrm{d}a_t |\Lcal_1 ]\\
    &=\E[\sum_{t=1}^H w_t(S_t) \int \{\hat c_t(S_t)-c_t(S_t)\}u_t(g_t)\{\bpol_t(g_t|S_t)-\hat \pi^b_t(g_t|S_t)\}\hat q_t(g_t,S_t)\mathrm{d}g_t  |\Lcal_1]+\\
  &+\E[\sum_{t=1}^H w_t(S_t)\{-\hat c^2_t(S_t)/c_t(S_t)+\hat c_t(S_t)+\hat c_t(S_t)-c_t(S_t) \}\int u_t(a_t)\hat \pi^b_t(a_t|S_t)\hat q_t(a_t,S_t)\mathrm{d}a_t |\Lcal_1 ]\\
     &=\E[\sum_{t=1}^H w_t(S_t)\int \{\hat c_t(S_t)-c_t(S_t)\}u_t(g_t)\{\bpol_t(g_t|S_t)-\hat \pi^b_t(g_t|S_t)\}\hat q_t(g_t,S_t)\mathrm{d}g_t |\Lcal_1 ]+\\
  &+\E[\sum_{t=1}^H w_t(S_t)\{-\hat c_t(S_t)+c_t(S_t)\}^2/c_t(S_t)\int u_t(a_t)\hat \pi^b_t(a_t|S_t)\hat q_t(a_t,S_t)\mathrm{d}a_t  |\Lcal_1]\\
  & \lessapprox \| \bpol_t(A_t|S_t)-\hat \pi^b_t(A_t|S_t)\|_2 \lessapprox \alpha^2_2.  
\end{align*}
This concludes that \cref{eq:f2_tmdp1} is $\alpha_1\beta+\alpha_2\beta+\alpha_1\alpha_2+\alpha^2_2$. Under the assumption for the convergence rates, this is equal to $\op(n^{-1/2})$.

\paragraph{Part 2:\cref{eq:f1_tmdp1} is $\op(1/\sqrt{n})$ }
~
Following \citet{KallusUehara2019}, this is proved if the following is proved:
\begin{align*}
\E[\{\phi(\hat w^{(1)},\hat \pi^{b(1)},\hat q^{(1)} ) -\phi(w, \pi^{b}, q^{} )\}^2 |\Lcal_1] =\op(1). 
\end{align*}
This is proved as in the Part 1 using the same decomposition. 
~
\paragraph{Final part }
~
\begin{align*}
    &\P_{\cu_1}[\phi(\hat w^{(1)},\hat \pi^{b(1)},\hat q^{(1)} )|\Lcal_1]+\E_{\cu_2}[\phi(\hat w^{(2)}, \hat \pi^{b(2)}, \hat q^{(2)} ) |\Lcal_2]\\
    &=\P_n[ \phi( w,\pi^b, q )]+\op(n^{-1/2}). 
\end{align*}
Then, CLT concludes the proof. 

\end{proof}

\begin{proof}[Proof of \cref{thm:partialdr_tmdp1}]

As we have ever seen, 
\begin{align*}
    \hat J_{\TI}-\P_n[ \phi( w^{\dagger},\pi^{b}, q^{\dagger} )]=\alpha_1\beta+\alpha_2\beta+\alpha_1\alpha_2+\alpha^2_2+\op(n^{-1/2}). 
\end{align*}
Under the assumption, the above is equal to $\op(1)$. 
In addition, the mean of $\P_n[ \phi( w^{\dagger},\pi^{b}, q^{\dagger} )]$ is $J$. Therefore, the statement holds from the law of large numbers. 

\end{proof}

\begin{proof}[Proof of \cref{thm:efficiency_tmdp2}]
For simplicity, we consider the case $K=2$. First, we prove
\begin{align}
   & \P_{\cu_1}[\phi(\hat w^{(1)},\hat \pi^{b(1)},\hat q^{(1)} ) |\Lcal_1]-\P_{\cu_1}[\phi(w, \pi^{b}, q ) |\Lcal_1] \nonumber \\
    &= \bG_{\cu_1}[ \phi(\hat w^{(1)},\hat \pi^{b(1)},\hat q^{(1)} )- \phi(w, \pi^{b}, q )  ]  \label{eq:f1_tmdp2}\\
    &+\E[\phi(\hat w^{(1)},\hat \pi^{b(1)},\hat q^{(1)} ) |\Lcal_1]-\E[\phi(w, \pi^{b}, q ) |\Lcal_1]   \label{eq:f2_tmdp2}\\
    &= \op(n^{-1/2}). \nonumber
\end{align}
~
\paragraph{ Part1: \cref{eq:f2_tmdp2} is $\op(n^{-1/2})$}
~
\begin{align*}
  \text{\cref{eq:f2_tmdp2}} &= \E[\phi(\hat w^{(1)},\hat \pi^{b(1)},\hat q^{(1)} ) |\Lcal_1]-\E[\phi(w, \pi^{b}, q ) |\Lcal_1] \\
    &=\E[\sum_{t=1}^H \{\hat \mu_t(S_t,A_t)-\mu_t(S_t,A_t)\}\{-\hat q_t(S_t,A_t)+q_t(S_t,A_t) \}\\ 
    &+\E[\sum_{t=1}^H \{\hat \mu_{t-1}(S_{t-1},A_{t-1})-\mu_{t-1}(S_{t-1},A_{t-1})\}\{-\hat q_t(S_t,\tau(S_t,A_t))+q_t(S_t,\tau(S_t,A_t)) \}  |\Lcal_1]\\
    &+\E[\sum_{t=1}^H  \mu_t(S_t,A_t)\{-\hat q_t(S_t,A_t)+q_t(S_t,A_t)  \}+\mu_{t-1}(S_{t-1},A_{t-1})\{\hat q_t(S_t,\tau(S_t,A_t))-q_t(S_t,\tau(S_t,A_t))\}   |\Lcal_1]\\
    &+\E[\sum_{t=1}^H \{\hat \mu_t(S_t,A_t)-\mu_t(S_t,A_t)\}\{R_t-q_t(S_t,A_t)+q_{t+1}(S_{t+1},\tau(S_{t+1},A_{t+1})) |\Lcal_1]\\
    &=\E[\sum_{t=1}^H \{\hat \mu_t(S_t,A_t)-\mu_t(S_t,A_t)\}\{-\hat q_t(S_t,A_t)+q_t(S_t,A_t) \}|\Lcal_1]\\
    &+\E[\sum_{t=1}^H \{\hat \mu_{t-1}(S_{t-1},A_{t-1})-\mu_{t-1}(S_{t-1},A_{t-1})\}\{-\hat q_t(S_t,\tau(S_t,A_t))+q_t(S_t,\tau(S_t,A_t)) \}  |\Lcal_1]\\
    &=\sum_{t=1}^H\|\hat \mu_t(S_t,A_t)-\mu_t(S_t,A_t)\|_2\| -\hat q_t(S_t,A_t)+q_t(S_t,A_t) \|_2\\
    &+\|\hat \mu_{t-1}(S_{t-1},A_{t-1})-\mu_{t-1}(S_{t-1},A_{t-1})\|_2\|-\hat q_t(S_t,\tau(S_t,A_t))+q_t(S_t,\tau(S_t,A_t)) \|_2 \\
    &=(\alpha_1+\alpha_2)\beta=\op(n^{-1/2}). 
\end{align*}
Here, we use 
\begin{align*}
    &\E[\mu_k(S_k,A_k) f(S_k,A_k)]=  \E[\prod_{t=1}^k\eta_t(S_{t},A_{t}) f(S_k,A_k)]=\E[(\prod_{t=1}^{k-1} \eta_t(S_{t},A_{t}))\eta_k (S_k,A_k)f(S_k,A_k)]\\
    &=\E[(\prod_{t=1}^{k-1} \eta_t(S_{t},A_{t})) f(S_k,\tau_k(A_k,S_k))]=\E[\mu_{k-1}(S_{k-1},A_{k-1}) f(S_k,\tau_k(A_k,S_k))], 
\end{align*}
and 
\begin{align*}
    \|\hat q(S,\tau(S,A))-q(S,\tau(S,A))\|^2_2&=\int \{\hat q(s,\tau(s,a))-q(s,\tau(s,a))\}^2\bpol(a|s)p(s)\mathrm{d}(s,a)\\
    &=\int \{\hat q(s,a)-q(s,a)\}^2\epol(a|s)p(s)\mathrm{d}(s,a)\\
    &\leq C'\|\hat q(S,A)- q(S,A)\|^2_2. 
\end{align*}

~
\paragraph{Part 2:\cref{eq:f1_tmdp2} is $\op(n^{-1/2})$ }
Following \citet{KallusUehara2019}, this is proved if the following is proved:
\begin{align*}
\E[\{\phi(\hat w^{(1)},\hat \pi^{b(1)},\hat q^{(1)} ) -\phi(w, \pi^{b}, q^{} )\}^2 |\Lcal_1] =\op(1). 
\end{align*}
This is proved as in the Part 1 using the same decomposition. 
~
\paragraph{Final part }
~
\begin{align*}
    &\P_{\cu_1}[\phi(\hat w^{(1)},\hat \pi^{b(1)},\hat q^{(1)} )|\Lcal_1]+\E_{\cu_2}[\phi(\hat w, \hat \pi^{b(2)}, \hat q^{(2)} ) |\Lcal_2]\\
    &=\P_n[ \phi( w,\pi^b, q )]+\op(n^{-1/2}). 
\end{align*}
Then, CLT concludes the proof. 

\end{proof}

\begin{proof}[Proof of \cref{thm:dr_estimatino2_tmdp}]

From the proof of \cref{thm:efficiency_tmdp2}, we have 
$$\hat J_{\TI1}=\P_n[ \phi( w^{\dagger},\pi^{b\dagger}, q^{\dagger} )]+(\alpha_1+\alpha_2)\beta+\op(n^{-1/2}).$$
From the law of large numbers, noting the expectation of $\P_n[ \phi( w^{\dagger},\pi^{b\dagger}, q^{\dagger} )]$ is $J$ under the assumption, the statement is concluded. 

\end{proof}

\subsection{Proofs for \cref{sec:nmdp}}

\begin{proof}[Proof of \cref{thm:nmdp_example1}]
~
\paragraph*{Calculation of derivatives under the nonparemtric NMDP.}

The entire regular parametric submodel corresponding an NMDP is 
\begin{align*}
    \{p_{\theta}(s_1)p_{\theta}(a_1|s_1)p_{\theta}(r_1|\cjj_{a_1})p_{\theta}(s_2|\cjj_{r_1})p_{\theta}(a_2|\cjj_{s_2})p_{\theta}(r_2|\cjj_{a_2})\cdots p_{\theta}(r_H|\cjj_{a_H})\}, 
\end{align*}
where it matches with the true pdf at $\theta=0$. The score function of the model is decomposed as 
\begin{align*}
    g(j_{r_H})&=\sum_{k=1}^{H} \nabla \log p_{\theta}(s_{k}|\cjj_{r_{k-1}}) +\sum_{k=1}^{H} \nabla \log p_{\theta}(a_{k}|\cjj_{s_k}) +\sum_{k=1}^{H} \nabla \log p_{\theta}(r_k|\cjj_{a_k})\\ 
    &=\sum_{k=1}^{H}g_{S_{k}|\cj_{r_{{k-1}}}}+\sum_{k=1}^{H}g_{A_{k}|\cj_{s_k}}+\sum_{k=1}^{H}g_{R_k|\cj_{a_k}}. 
\end{align*}
Then, we have 
\begin{align*}
     \nabla_{\theta} \mathrm{E}_{\pi^{e}}\left [\sum_{t=1}^{H}r_t \right]= \E[\{-J+\sum_{c=1}^{H}\lambda_c\{r_c-v_c(\cj_{s_c})\}+\lambda_{c-1}v_c(\cj_{s_c})\} g(\mathcal{J}_{r_T})]. 
\end{align*}
This is proved by 
\begin{align*}
& \nabla_{\theta} \mathrm{E}_{\pi^{e}}\left [\sum_{t=1}^{H}r_t \right] =\nabla_{\theta} \left[\int \sum_{t=1}^{H}r_t \left\{\prod_{k=1}^{H}p_{\theta} (s_k|\cjj_{r_{k-1}})\pi^{e}(a_k|\cjj_{s_k};\theta)p_{\theta}(r_k|\cjj_{a_k}) \right\}\mathrm{d}\mu(\cjj_{r_H})\right]   \\
    &=\sum_{c=1}^{H} \{\E_{\pi^{e}}[\{\E_{\pi^{e}}(r_c|s_1)-\E_{\pi^{e}}(r_c)\}g_{s_1}]+\E_{\pi^{e}}[\{r_c-\E_{\pi^{e}}(R_c|\cj_{a_c})\} g_{r_c|\cj_{a_c}}] \\
    &+\E_{\pi^{e}}\left[\left(\E_{\pi^{e}}\left [\sum_{t=c+1}^{H} r_t|\cj_{s_{c+1}}\right]-\E_{\pi^{e}}\left [\sum_{t=c+1}^{H} r_t|\cj_{a_c}\right ]\right)g_{S_{c+1}|\cj_{r_c}}\right] \\
    &+\E[\lambda_c\{\E_{\pi^{e}}[\sum_{t=c}^{H}r_t \mid \cj_{a_c}]-\E_{\pi^{e}}[\sum_{t=c}^{H}r_t \mid \cj_{s_c}]\} g_{A_c\mid \cj_{s_c}}]  \}. 
\end{align*}  
Note that except for the third term, the proof is the same as Theorem 2. The third term is calculated by 
{\small
\begin{align*}
 &\sum_{m=1}^H\int \sum_{t=1}^{H} \left\{r_t\prod_{k=1,k\neq m}^{t}p_{\theta} (s_k|\cjj_{r_{k-1}}) \pi^{e}(a_k|\cjj_{s_k};\theta)p_{\theta}(r_k|\cjj_{a_k}) \right\}\times \\
 &\,\,\,\,\,\epol_m\left\{g_{A_m|\cj_{s_m}}-\left[\int g_{A_m|H_{S_m}}\epol(a_m\mid \cjj_{s_m})\mathrm{d}(a_m)\right]\right \}\mathrm{d}\mu(j_{r_H}) \\
 &=\sum_{m=1}^H \E_{\epol}[\sum_{t=m}^H r_t g_{A_m | \cj_{s_m}}-\sum_{t=m}^H r_t \E_{\epol}[g_{A_m | H_{s_m}}\mid  \cjj_{s_m}]] \\
  &=\sum_{m=1}^H \E_{\epol}[\E_{\epol}[\sum_{t=m}^H r_t\mid \cjj_{A_m}] g_{A_m | \cj_{s_m}}-\E_{\epol}[\sum_{t=m}^H r_t\mid \cjj_{s_m} ] \E_{\epol}[g_{A_m | H_{s_m}}\mid  \cjj_{s_m}]] \\
  &=\E[\lambda_c\{\E_{\pi^{e}}[\sum_{t=c}^{H}r_t \mid \cj_{a_c}]-\E_{\pi^{e}}[\sum_{t=c}^{H}r_t \mid \cj_{s_c}]\} g_{A_c\mid \cj_{s_c}}]  . 
\end{align*}
}
Then, $\nabla_{\theta} \mathrm{E}_{\pi^{e}}\left [\sum_{t=1}^{H}r_t \right]$ is equal to 
\begin{align*}
    &\sum_{c=1}^{H}\{\E_{\pi^{e}}[\{\E_{\pi^{e}}(r_c|s_1)-\E_{\pi^{e}}(r_c)\}g(\mathcal{J}_{r_{H}})]+\E_{\pi^{e}}[\{r_c-\E_{\pi^{e}}(r_c|\cj_{a_c})\} g(\mathcal{J}_{r_H})] \\ 
    &+\E_{\pi^{e}}\left[\left(\E_{\pi^{e}}\left [\sum_{t=c+1}^{H} r_t|\cj_{s_c+1}\right]-\E_{\pi^{e}}\left [\sum_{t=c+1}^{H} r_t|\cj_{a_c}\right ]\right)g(\mathcal{J}_{r_H})\right]\} \\
     &+\E[\lambda_c\{\E[\sum_{t=c}^{H}r_t \mid \cj_{a_c}]-\E[\sum_{t=c}^{H}r_t \mid \cj_{s_c}]\} g(\mathcal{J}_{r_H})]  \}\\
    &= \E \left(\left[-J+\sum_{c=1}^{H}\left \{\lambda_{c}r_c-\lambda_{c}\sum_{t=c}^{H}\E_{\pi_e}(r_t|\cj_{a_c})+\lambda_{c-1}\sum_{t=c}^{H}\E_{\pi_e}(r_t|\cj_{s_c})\right \}\right]g(\mathcal{J}_{r_H})\right)+\\
    &+\E[\sum_{c=1}^{H}\{\lambda_{c}\sum_{t=c}^{H}\E_{\pi_e}(r_t|\cj_{a_c})-\lambda_{c}\E_{\pi_e}[\sum_{t=c}^{H}r_t \mid \cj_{s_c}] \}g(\mathcal{J}_{r_H})]\\
    &=\E[\{-J+\sum_{c=1}^{H}\lambda_c\{r_c-v_c(\cj_{s_c})\}+\lambda_{c-1}v_c(\cj_{s_c})\} g(\mathcal{J}_{r_H})]. 
\end{align*}

This concludes that the following is a derivative: 
\begin{align}
\label{eq:gradient_example1_nmdp}
    -J+\sum_{c=1}^{H}\lambda_c\{r_c-v_c(\cjj_{s_c})\}+\lambda_{c-1}v_c(\cjj_{s_c}).
\end{align}
This concludes that $\phi$ is actually the EIF since the derivativee is unique. 

\paragraph*{Calculation of the efficiency bound}

From the law of total variance, the efficiency bound is 
\begin{align*}
    &\var\left[ \sum_{c=1}^H\lambda_c\{R_c-v_{c}(\cj_{S_c})\}+\lambda_{c-1}v_{c}(\cj_{S_c}) \right]\\
    &=\sum_{t=0}^H  \E[\var[\E[  \sum_{c=1}^H\lambda_c\{R_c+v_{c+1}(\cj_{S_{c+1}})-v_{c}(\cj_{S_c})\} |\cj_{S_{t+1}}]|\cj_{S_{t}}] ] \\
    &=\sum_{t=0}^{H}\E[\var[\lambda_t\{R_t+v_{t+1}(\cj_{S_{t+1}})\}\mid \cj_{S_t}]]=\sum_{t=0}^{H}\E[\lambda^2_t\var[R_t+v_{t+1}(\cj_{S_{t+1}}) \mid \cj_{S_t}]].
\end{align*}

\end{proof}

\begin{proof}[Proof of \cref{thm:nmdp_variance}]
\begin{align*}
   &\sum_{t=0}^{H}\mathrm{E}_{\pi_b}\left[\lambda^{2}_{t-1}\mathrm{var}\left\{\eta_t \{r_{t}+v_{t+1}(\cjj_{s_{t+1}})\}\mid \cjj_{s_{t}}\right \}\right] \\
   & =   \sum_{t=0}^{H}\mathrm{E}_{\pi_e}\left[\lambda_{t-1}\mathrm{var}\left\{\eta_t \{r_{t}+v_{t+1}(\cjj_{s_{t+1}})\}\mid \cjj_{s_{t}}\right \}\right]\\
   &\geq
   \sum_{t=0}^{H}\exp\mathrm{E}_{\epol}\left[\log(\lambda_{t-1}\mathrm{var}\left\{\eta_t(r_{t}+v_{t+1}(\cjj_{s_{t+1}})\mid \cjj_{s_{t}}\right \})\right]
   \\
   & \geq \sum_{t=0}^{H}\exp(\mathrm{E}_{\epol}\left[\log(\lambda_{t-1})\right]+\mathrm{E}_{\epol}\left[\log(\mathrm{var}\left\{(\eta_t\{r_{t}+v_{t+1}(\cjj_{s_{t+1}})\}\mid \cjj_{s_{t}}\right \})\right]) \\
   &\geq
  \sum_{t=0}^{H}\exp((t-1)\log C_{\min}+\log V_{\min}^2) \geq V_{\min}^2C_{\min}^{H-1}.
\end{align*}
\end{proof}

\begin{proof}[Proof of \cref{thm:nmdp_example2}]
Almost the same as the proof of \cref{thm:mdp_example2}
\end{proof}

\begin{proof}[Proof of \cref{thm:nmdp_variance2}]
\begin{align*}
   &\sum_{t=0}^{H}\E[\lambda^2_t \var[r_t+q_{t+1}(\cjj_{s_{t+1}},\tau_{t+1}(\cjj_{a_{t+1}}))\mid \cjj_{a_t}]] \\
   & =   \sum_{t=0}^{H}\E_{\epol}[\lambda_t \var[r_t+q_{t+1}(\cjj_{s_{t+1}},\tau_{t+1}(\cjj_{a_{t+1}}))\mid \cjj_{a_t}]]\\
   &\geq
 \sum_{t=0}^{H}\exp\E_{\epol}[\log(\lambda_t \var[r_t+q_{t+1}(\cjj_{s_{t+1}},\tau_{t+1}(\cjj_{a_{t+1}}))\mid \cjj_{a_t}])]
   \\
   & \geq \sum_{t=0}^{H}\exp(\E_{\epol}[\log(\lambda_t)]+\E_{\epol}[\log( \var[r_t+q_{t+1}(\cjj_{s_{t+1}},\tau_{t+1}(\cjj_{a_{t+1}}))\mid \cjj_{a_t}])]) \\
   &\geq
  \sum_{t=0}^{H}\exp((t)\log C_{\min}+\log V_{\min}^2)\geq V_{\min}^2C_{\min}^{H}.
\end{align*}
\end{proof}

\subsection{Proofs for \cref{sec:mdp}}

\begin{proof}[Proof of \cref{thm:mdpmdp_example1}]

In this section, we define $c_T(\gamma)=\sum_{t=1}^T \gamma^{t-1}$. Here, the entire regular parametric submodel is 
\begin{align*}
    \{p_{\theta}(a\mid s)p_{\theta}(r|a,s)p_{\theta}(s'|s,a)p_{\theta}(r|s,a)\},
\end{align*}
where it matches with the true pdf at $\theta=0$.The score function of the model $\cm_{\mathrm{MDP}}$ is decomposed as  
\begin{align*}
    g(o)=\log p_{\theta}(s)+\log p_{\theta}(a|s)+\log p_{\theta}(s'|s,a)+\log p_{\theta}(r|s,a)=g_{S}+g_{A|S}+g_{R|S,A}+g_{S'|S,A}. 
\end{align*}
Here, the corresponding tangent space is 
\begin{align*}
   &\Tcal_1 \times  \Tcal_2 \times  \Tcal_3 \times \Tcal_4, \,\\
   \Tcal_1 &=\braces{f(s);\E[f]=0,\var[f]<\infty }\,,\Tcal_2=\braces{f(s,a);\E[f(s,a)|s]=0,\var[f(s,a)|s]<\infty},\\
   \Tcal_3 &=\braces{f(s,a,s');\E[f(s,a,s')|s,a]=0,\var[f(s,a,s')|s,a]<\infty},\\
   \Tcal_4 &=\braces{f(s,a,r);\E[f(s,a,r)|s,a]=0,\var[f(s,a,r)|s,a]<\infty}. 
\end{align*}

We calculate the gradient of the target functional $J(\epol)$ w.r.t. the model $\cm_{\mathrm{MDP}}$. Since 
\begin{align*}
  J(\epol)&=\int r p_{e,\gamma}^{(\infty)}(s)\epol(a\mid s)p(r\mid s,a)\mathrm{d}(s,a,r)\\
  &=\lim_{T\to \infty}\int c_{T}(\gamma)\sum_{t=1}^{H}\int \gamma^{t-1} r_tp_{\theta}(r_t|s_t,a_t)\left\{\prod_{k=1}^t \epol(a_k|s_k;\theta)p_{\theta}(s_{k+1}|s_k,a_k)\right\}p^{(1)}_{e}(s_1)\mathrm{d}(h_{s_{t+1}})\\
  &=\lim_{T\to \infty}\int c_{T}(\gamma)\sum_{t=1}^{H} \E_{\epol}[\gamma^{t-1} r_t],
\end{align*}
what we need is deriving the gradient $\phi$ satisfying 
\begin{align*}
    \nabla J(\epol)=\E_{p_{b}}[\phi(s,a,r,s')g(s,a,r,s')]. 
\end{align*}
This is done as follows:  $\nabla J(\epol)=$
{\small 
\begin{align*}
       &\lim_{T\to \infty}\int c_{T}(\gamma)\sum_{t=1}^{H}\int \gamma^{t-1} r_t \nabla p_{\theta}(r_t|s_t,a_t)\braces{\prod_{k=1}^t \epol(a_k|s_k;\theta)p_{\theta}(s_{k+1}|s_k,a_k)}p^{(1)}_{e}(s_1)\mathrm{d}(\cjj_{s_{t+1}})\\
        &+\lim_{T\to \infty}\int c_{T}(\gamma)\sum_{t=1}^{H}\int \gamma^{t-1} r_tp_{\theta}(r_t|s_t,a_t) \braces{\prod_{k=1}^t \epol(a_k|s_k;\theta)\nabla p_{\theta}(s_{k+1}|s_k,a_k)}p^{(1)}_{e}(s_1)\mathrm{d}(\cjj_{s_{t+1}})\\
        &+\lim_{T\to \infty}\int c_{T}(\gamma)\sum_{t=1}^{H}\int \gamma^{t-1} r_tp_{\theta}(r_t|s_t,a_t)\braces{\prod_{k=1}^t \nabla \epol(a_k|s_k;\theta)p_{\theta}(s_{k+1}|s_k,a_k)}p^{(1)}_{e}(s_1)\mathrm{d}(\cjj_{s_{t+1}}). 
\end{align*}
}
\paragraph{First term}

This is equal to 
\begin{align*}
    &\lim_{T\to \infty}\int c_{T}(\gamma)\sum_{t=1}^{H} \E_{\epol}[\gamma^{t-1} r_t g_{R|S,A}(r_t|s_t,a_t)]=\E_{p^{(\infty)}_{e,\gamma}}[r g_{R|S,A}(r|s,a)]\\
    &=\E_{p^{(\infty)}_{e,\gamma}}[\{r-\E[r|s,a]\}g_{R|S,A}(r|s,a)]=\E_{p^{(\infty)}_{e,\gamma}}[\{r-\E_{p^{(\infty)}_{e,\gamma}}[r|s,a]\}g(s,a,r,s')].
\end{align*}

\paragraph{Second term}

This is equal to 
\begin{align*}
     &\lim_{T\to \infty}\gamma \int c_{T}(\gamma)\sum_{c=1}^{H}\gamma^c \sum_{t=c+1}^{H}\E_{\epol}[\gamma^{t-c-2}r_t g_{S'|S,A}(s_{t+1}|s_t,a_t) ]=\gamma \E_{p^{(\infty)}_{e,\gamma}}[v(s')g_{S'|S,A}(s'|s,a)]\\
     &=\gamma \E_{p^{(\infty)}_{e,\gamma}}[v(s')g_{S'|S,A}(s'|s,a)]=\gamma \E_{p^{(\infty)}_{e,\gamma}}[\{v(s')-\E_{p^{(\infty)}_{e,\gamma}}[v(s')|s,a]\}g_{S'|S,A}(s'|s,a)]\\
     &=\gamma \E_{p^{(\infty)}_{e,\gamma}}[\{v(s')-\E_{p^{(\infty)}_{e,\gamma}}[v(s')|s,a]\}g(s,a,r,s')]. 
\end{align*}

\paragraph{Third term}

This is equal to 
\begin{align*}
     &\lim_{T\to \infty}\int c_{T}(\gamma)\sum_{c=1}^{H}\gamma^c \sum_{t=c+1}^{H}\E_{\epol}[\gamma^{t-c-1}r_t \{g_{A|S}(a_t|s_t)-\E_{\epol}[g_{A|S}(a_t|s_t)|s_t]\}]\\
     &=\E_{p^{(\infty)}_{e,\gamma}}[\{q(s,a)-v(s)\}g_{S'|S,A}(s'|s,a)]=\E_{p^{(\infty)}_{e,\gamma}}[\{q(s,a)-v(s)\}g(s,a,r,s')]. 
\end{align*}

In summary, 
\begin{align*}
     \nabla J(\epol)&= \E_{p^{(\infty)}_{e,\gamma}}\left[\left\{r-\E_{p^{(\infty)}_{e,\gamma}}[r|s,a]+\gamma v(s')-\gamma \E[v(s')|s,a]+q(s,a)-v(s)\right\} g(s,a,r,s')\right]\\ 
     &=\E_{p^{(\infty)}_{e,\gamma}}[\{r+\gamma v(s')-v(s)\} g(s,a,r,s') ]\\
    &=\E_{p_{b}}[\mu^{*}(s,a)\{r+\gamma v(s')-v(s)\}g(s,a,r,s') ]. 
\end{align*}
Thus, the following function:
\begin{align*}
    \mu^{*}(s,a)\{r+\gamma v(s')-v(s)\}. 
\end{align*}
is a gradient. Besides, since it belongs to a tangent space, this is the EIF. The efficiency bound is 
\begin{align*}
    \E [w^{*2}(S)\var[\eta(S,A)\{R+\gamma v(S')\}|S]]. 
\end{align*}
\end{proof}

\begin{proof}[Proof of \cref{thm:mdp_estimation1}]
For simplicity, we consider the case $K=2$. Here, we have 
\begin{align}
   & \P_{\cu_1}[\phi(\hat w^{*(1)},\hat \pi^{b(1)},\hat q^{(1)} ) |\Lcal_1]-\P_{\cu_1}[\phi(w^{*}, \pi^{b}, q^{} ) |\Lcal_1] \nonumber \\
    &= \bG_{\cu_1}[ \phi(\hat w^{*(1)},\hat \pi^{b(1)},\hat q^{(1)} )- \phi(w^{*}, \pi^{b}, q )  ]  \label{eq:f1_mdp1}\\
    &+\E[\phi(\hat w^{*(1)},\hat \pi^{b(1)},\hat q^{(1)} ) |\Lcal_1]-\E[\phi(w^{*}, \pi^{b}, q^{} ) |\Lcal_1]   \label{eq:f2_mdp1}\\
    &= \op(n^{-1/2}). \nonumber
\end{align}
Then, the proof is immediately concluded from CLT. For the rest of the proof, we prove \cref{eq:f2_mdp1} is $\op(n^{-1/2})$. The part \cref{eq:f1_mdp1} is $\op(n^{-1/2})$ is similarly proved from the same decomposition.  

We have 
\begin{align*}
    &\E[\hat w^{*}(S)\hat \eta(S,A)\{R-\hat q(S,A)+\gamma \hat v(S')\}]+ \E_{p^{(1)}_e}[\hat v(S)]-J\\
    &=\E[ \{ \hat w^{*}(S)\hat\eta(S,A) -w^{*}(S)\eta(S,A)\}\{R-q(S,A)+\gamma v(S')\}]\\
    &+\E[w^{*}(s)\eta(S,A)\{q(S,A)-\hat q(S,A)\}+w^{*}(s)\{\gamma \hat v(S')-\gamma v(S')\}]+(1-\gamma)\E_{p^{(1)}_e}[\hat v(s_1)-v(s_1)]\\
    &+\E[ \{\hat w^{*}(S)\hat \eta(S,A)-w^{*}(S)\eta(S,A)\}\{q(S,A)-\hat q(S,A)+\gamma  \hat v(S')-\gamma v(S')\}] \\
    &\lessapprox \E[w^{*}(S)\eta(S,A)\{-\hat q(S,A)+\gamma \hat v(S')\}]+(1-\gamma) \E_{p^{(1)}_e}[\hat v(s_1)] \\
    &\,\,\,\,+\|\hat q(S,A)-\hat q(S,A)\|_2\|\hat \mu(S,A)-\hat \mu(S,A)\|_2\\
&\lessapprox \E[w^{*}(S)\eta(S,A)\{-\hat q(S,A)+\gamma \hat v(S')\}]+(1-\gamma) \E_{p^{(1)}_e}[\hat v(s_1)]+\alpha_1\alpha_2+\alpha_1\beta+\alpha_2\beta+\alpha^2_2. 
\end{align*}
Then, noting what we want to control is $\E[\hat w^{*}(S)\hat \eta(S,A)\{R-\hat v(S)+\gamma \hat v(S')\}]+ \E_{p^{(1)}_e}[\hat v(S)]$, we have 
\begin{align*}
    &\E[\hat w^{*}(S)\hat \eta(S,A)\{R-\hat v(S)+\gamma \hat v(S')\}]+ \E_{p^{(1)}_e}[\hat v(S)]-J\\
    &= \E[w^{*}(S)\eta(S,A)\{-\hat q(S,A)+\gamma \hat v(S')\}]+(1-\gamma) \E_{p^{(1)}_e}[\hat v(S)]\\
    &+\E[\hat w^{*}(S)\hat \eta(S,A)\{\hat q(S,A)-\hat v(S)\} ]+\alpha_1\alpha_2+\alpha_1\beta+\alpha_2\beta+\alpha^2_2\\
    &=\E[w^{*}(S)\eta(S,A)\{-\hat q(S,A)+\gamma v(S')\}]+\E[w^{*}(S)\hat \eta(S,A)\{\hat q(S,A)-\hat v(S)\} ]\\
    &+\E[-\gamma w^{*}(S)\eta(S,A)\hat v(S') +w^{*}(S)\hat v(S)]+\alpha_1\alpha_2+\alpha_1\beta+\alpha_2\beta+\alpha^2_2\\
    &=\E[w^{*}(S)\hat \eta(S,A)\{\hat q(S,A)-\hat v(S)\} ]+\E[w^{*}(S)\eta(S,A)\{-\hat q(S,A)+\hat v(S)\}]+\\
    &\alpha_1\alpha_2+\alpha_1\beta+\alpha_2\beta+\alpha^2_2.
\end{align*}
The last term is equal to $\alpha^2_2$ as follows:
\begin{align*}
    &\E[w^{*}(S)\hat \eta(S,A)\{\hat q(S,A)-\hat v(S)\} ]+\E[w^{*}(S)\eta(S,A)\{-\hat q(S,A)+\hat v(S)\}]\\
    &=\E\left[w^{*}(S)\hat c(S)u(A) \hat \pi^b(A|S)\{-\hat v(S)+\hat q(S,A)\} \right]+\E[w^{*}(S)\int \{\hat \pi^e(a|S)-\pi^e(a|S)\}\hat q(a,S)\mathrm{d}a]\\
  &=\E\left[w^{*}(S)\int \hat c(S)u(g)\bpol(g|s)\{-\hat c(S)\int u(a)\hat q(a,S)\hat \pi^b(a|S)\mathrm{d}a +\hat q(g,S)\}\mathrm{d}g \right]\\
  &+\E[w^{*}(S)\int u(a)\{\hat c(S)\hat \pi^b(a|S) -c(S)\pi^b(a|S) \}\hat q(a,S)\mathrm{d}a]\\
  &=\E[w^{*}(S)\int \{\hat c(S)-c(S)\}u(g)\bpol(g|s)\hat q(g,S)\mathrm{d}g  ]+\\
  &+\E[w^{*}(S)\{-\hat c^2(S)/c(S)+\hat c(S) \}\int u(a)\hat \pi^b(a|S)\hat q(a,S)\mathrm{d}a  ]\\
    &=\E[w^{*}(S)\int \{\hat c(S)-c(S)\}u(g)\{\bpol(g|s)-\hat \pi^b(g|S)\}\hat q(g,S)\mathrm{d}g  ]+\\
  &+\E[w^{*}(S)\{-\hat c^2(S)/c(S)+\hat c(S)+\hat c(S)-c(S) \}\int u(a)\hat \pi^b(a|S)\hat q(a,S)\mathrm{d}a  ]\\
     &=\E[w^{*}(S)\int \{\hat c(S)-c(S)\}u(g)\{\bpol(g|S)-\hat \pi^b(g|S)\}\hat q(g,S)\mathrm{d}g  ]+\\
  &+\E[w^{*}(S)\{-\hat c(S)+c(S)\}^2/c(S)\int u(a)\hat \pi^b(a|S)\hat q(a,S)\mathrm{d}a  ]\\
  &\lessapprox \|\hat \pi^b(A|S)-\bpol(A|S) \|^2_2=\alpha^2_2.
\end{align*}
In summary,
\begin{align*}
    \E[\phi(\hat w^{*(1)},\hat \pi^{b(1)},\hat q^{(1)} ) |\Lcal_1]-\E[\phi(w^{*}, \pi^{b}, q^{} ) |\Lcal_1]=\alpha_1\alpha_2+\alpha_1\beta+\alpha_2\beta+\alpha^2_2. 
\end{align*}
Under the assumption regarding convergence rates, this is equal to $\op(n^{-1/2})$. 

\end{proof}

\begin{proof}[Proof of \cref{thm:dr_estimatino1_mdp}]
This is immediately concluded from 
\begin{align*}
    \E[\phi(\hat w^{*(1)},\hat \pi^{b(1)},\hat q^{(1)} ) |\Lcal_1]-\E[\phi(w^{*\dagger}, \pi^{b\dagger}, q^{\dagger} ) |\Lcal_1]=\alpha_1\alpha_2+\alpha_1\beta+\alpha_2\beta+\alpha^2_2, 
\end{align*}
which is obtained in the proof of \cref{thm:mdp_estimation1}. 
\end{proof}

\begin{proof}[Proof of \cref{thm:mdpmdp_example2}]
This is similarly proved as in \cref{thm:mdpmdp_example1} by some calculation.  
\end{proof}

\begin{proof}[Proof of \cref{thm:mdp_estimation2}]
Here, we have 
\begin{align}
   & \P_{\cu_1}[\phi(\hat w^{*(1)},\hat \pi^{b(1)},\hat q^{(1)} ) |\Lcal_1]-\P_{\cu_1}[\phi(w^{*}, \pi^{b}, q^{} ) |\Lcal_1] \nonumber \\
    &= \bG_{\cu_1}[ \phi(\hat w^{*(1)},\hat \pi^{b(1)},\hat q^{(1)} )- \phi(w^{*}, \pi^{b}, q^{} )  ]  \label{eq:f1_mdp}\\
    &+\E[\phi(\hat w^{*(1)},\hat \pi^{b(1)},\hat q^{(1)} ) |\Lcal_1]-\E[\phi(w^{*}, \pi^{b}, q^{} ) |\Lcal_1]   \label{eq:f2_mdp}\\
    &= \op(n^{-1/2}). \nonumber
\end{align}
Then, the proof is immediately concluded from CLT. It remains to bound \cref{eq:f2_mdp} is $\op(n^{-1/2})$. The part that \cref{eq:f1_mdp} is $\op(n^{-1/2})$ is similarly proved from the same decomposition. We have 
\begin{align}
    &\E[\hat w^{*}(S)\hat \eta(S,A)\{R-\hat q(S,A)+\gamma \hat q(S',\tau(A',S'))\}]+(1-\gamma)\E_{p^{(1)}_e}[\hat v(s_1)]-J  \nonumber\\
    &=\E[ \{ \hat w^{*}(S)\hat \eta(S,A) -w^{*}(S)\eta(S,A)\}\{R-q(S,A)+\gamma q(S',\tau(A',S'))\}]  \label{eq:mdp_est2}\\
    &+\E[w^{*}(S)\eta(S,A)\{q(S,A)-\hat q(S,A)\}+w^{*}(S)\{\gamma \hat q(S,\tau(S,A))-\gamma q(S,\tau(S,A))\}] \nonumber \\
    &+(1-\gamma)\E_{p^{(1)}_e}[\hat v(s_1)-v(s_1)] \label{eq:mdp_est3} \\
    &+\E[ \{ \hat w^{*}(S)\hat \eta(S,A) -w^{*}(S)\eta(S,A)\}\{q(S,A)-\hat q(S,A)+\gamma \hat q(S',\tau(A',S'))-\gamma q(S',\tau(A',S'))\}] \nonumber \\
    &\lessapprox \|\hat w^{*}(S)\hat \eta(S,A)-w^{*}(S)\eta(S,A)\|_2\|\hat q(S,A)-q(S,A)\|_2=(\alpha_1+\alpha_2)\beta.  \nonumber
\end{align}
Here, we use \cref{eq:mdp_est2} and \cref{eq:mdp_est3}  are $0$. 
\end{proof}
\begin{proof}[Proof of \cref{thm:dr_estimatino2_mdp}]
This is immediately concluded from the proof of \cref{thm:mdp_estimation2} as in Corollary 16 \citep{KallusUehara2019}. 
\end{proof}

\begin{proof}[Proof of \cref{thm:dr_value}]
Immediately, concluded from \cref{thm:jv_mse}. 
\end{proof}

\begin{proof}[Proof of \cref{thm:jv_mse}]

Here, we have 
\begin{align}
   & \P_{\cu_1}[\phi(\hat w^{*(1)},\hat v^{(1)} ) |\Lcal_1]-\P_{\cu_1}[\phi(w^{*}, v) |\Lcal_1] \nonumber \\
    &= \bG_{\cu_1}[ \phi(\hat w^{*(1)},\hat v^{(1)} )- \phi(w^{*},  v)  ]  \label{eq:f1_mdp_v}\\
    &+\E[\phi(\hat w^{*(1)},\hat v^{(1)} ) |\Lcal_1]-\E[\phi(w^{*},  v)  |\Lcal_1]   \label{eq:f2_mdp_v}\\
    &= \op(n^{-1/2}). \nonumber
\end{align}
Then, the proof is immediately concluded from CLT. For the rest of the proof, we prove \cref{eq:f2_mdp_v} is $\op(n^{-1/2})$. The part \cref{eq:f1_mdp_v} is $\op(n^{-1/2})$ is similarly proved from the same decomposition. We have 
\begin{align}
    &\E[\hat w^{*}(S)\eta(S,A)\{R-\hat v(S)+\gamma \hat v(S')\}]+(1-\gamma)\E_{p^{(1)}_e}[\hat v(s_1)]-J  \nonumber \\
    &=\E[ \{ \hat w^{*}(S)\eta(S,A) -w^{*}(S)\eta(S,A)\}\{R-v(S)+\gamma v(S')\}] \label{eq:jv_mse1}\\
    &+\E[w^{*}(S)\eta(S,A)\{v(S)-\hat v(S)\}+w^{*}(S)\{\gamma \hat v(S)-\gamma v(S')\}]  +(1-\gamma)\E_{p^{(1)}_e}[\hat v(s_1)-v(s_1)] \label{eq:jv_mse2}\\
    &+\E[ \{ \hat w^{*}(S)\eta(S,A) -w^{*}(S)\eta(S,A)\}\{v(S)-\hat v(S)+\gamma \hat v(S')-\gamma v(S')\}]  \nonumber \\
    &\lessapprox \|\hat w^{*}(S)\eta(S,A)-w^{*}(S)\eta(S,A)\|_2\|\hat v(S)-v(S)\|_2=\alpha_1\beta=\op(n^{-1/2}).  \nonumber
\end{align}
Here, we use the following fact:
\begin{align*}
    \E[\eta(S,A)\{R-v(S)+\gamma v(S')\}]=\E[\E[\eta(S,A)\{R+\gamma v(S')\}|S]-v(S)]=0, 
\end{align*}
and we derived that \cref{eq:jv_mse2} and  \cref{eq:jv_mse1} are $0$.  

\begin{remark}
In the finite horizon case, we also have a double robustness and efficiency. The argument is as follows. Here, we have 
\begin{align}
   & \P_{\cu_1}[\phi(\hat w^{(1)},\hat v^{(1)} ) |\Lcal_1]-\P_{\cu_1}[\phi(w, v) |\Lcal_1] \nonumber \\
    &= \bG_{\cu_1}[ \phi(\hat w^{(1)},\hat v^{(1)} )- \phi(w, v )  ]  \label{eq:f1_other}\\
    &+\E[\phi(\hat w^{(1)},\hat v^{(1)} ) |\Lcal_1]-\E[\phi(w, v ) |\Lcal_1]   \label{eq:f2_other}\\
    &= \op(n^{-1/2}). \nonumber
\end{align}
We prove \cref{eq:f2_other} is $\op(n^{-1/2})$. The fact \cref{eq:f1_other} is $\op(n^{-1/2})$ is similarly proved as follows. Then, the final statement is concluded from CLT. The proof is as follows:
\begin{align*}
    &\E[ \sum_{t=1}^H \hat w_t(S_t)\eta(S_t,A_t)\{R_t-\hat v_t(S_t)\}+\hat w_{t-1}(S_{t-1})\eta(S_{t-1},A_{t-1})\hat v_t(S_t) ]\\
    &=  \E[ \sum_{t=1}^H \{\hat w_t(S_t)-w_t(S_t)\}\eta_t(S_t,A_t)\{-\hat v_t(S_t)+v_t(S_t)\} \\
    &+  \E[ \sum_{t=1}^H \{\hat w_{t-1}(S_{t-1})-w_{t-1}(S_{t-1})\}\eta_{t-1}(S_{t-1},A_{t-1})\{-\hat v_t(S_t)+v_t(S_t)\}  ]\\
    &+ \E[ \sum_{t=1}^H  w_t(S_t)\{-\hat v_t(S_t)+v_t(S_t)\}+ w_{t-1}(S_{t-1})\{\hat v_{t}(S_t)-v_{t}(S_t) \} ]\\
    &+ \E[ \sum_{t=1}^H \{\hat w_t(S_t)-w_t(S_t)\} \eta_t(S_t,A_t)\{R_t-v_t(S_t)+v_{t+1}(S_{t+1})\} ] \\
    &\lessapprox  \sum_{t=1}^H \|\hat v_t(S_t)-v_t(S_t)\|_2\|\hat w_t(S_t)-w_t(S_t) \|_2=\alpha_1\beta. 
\end{align*}
\end{remark}

\end{proof}

\section{Properties of the estimator of \citet{tang2019harnessing}}\label{sec:tangappendix}

In this section we consider a sample-splitting version of the estimator of \citet{tang2019harnessing} in the case of a pre-specified evaluation policy and establish its (inefficient) asymptotic behavior and double robustness.
Given nuisance estimators $\hat v^{(k)}(s),\hat w^{*(k)}(s)$,
the estimator we consider is given by replacing \cref{eq:central} in \cref{alg:tilting} with
\begin{align*}
 \hat J_k & =\frac{1}{|I_k|}\sum_{i\in I_k}[\hat w^{*(k)}(S^{(i)})\eta(S^{(i)},A^{(i)})\{R^{(i)}+\gamma \hat v^{(k)}(S^{'(i)})-\hat v^{(k)}(S^{(i)})\}]\\
 &+(1-\gamma)\E_{p^{(1)}_{e}(s_1)}[\hat v^{(k)}(s_1)].
\end{align*}
Here, we assume the behavior policy is known. We denote this estimator as $\hat J_{\mathrm{V2}}$. We can similarly consider an analogous estimator, $\hat J_{\mathrm{V1}}$, for the finite horizon case, where, given nuisance estimators $\hat v^{(k)}_t(s),\hat w^{(k)}_t(s)$, we use
\begin{align*}
 \hat J_k & =\frac{1}{|I_k|}\sum_{i\in I_k}\left [\hat v^{(k)}_1(S^{(i)}_1)+\sum_{t=1}^H\hat w^{(k)}_t(S^{(i)}_t)\eta_t(S^{(i)}_t,A^{(i)}_t)\{R^{(i)}_t+\hat v^{(k)}_{t+1}(S^{(i)}_{t+1})-\hat v^{(k)}_t(S^{(i)}_t)\}\right].
\end{align*}
Importantly, these estimators allow direct estimation of the value functions. In this sense, they are different from $\hat J_{\TI1},\hat J_{\TI2}$, where the construction of $\hat v(s)$ is restricted to $\hat c(s)\int u(a)\hat \pi^b(a|s)\hat q(s,a)\mathrm{d}a,\,\hat c(s)=1/\int u(a)\hat \pi^b(a|s)\mathrm{d}a$. 

The estimator $\hat J_{\mathrm{V2}}$ is doubly robust if the behavior policy is known in the sense that it is consistent as long as the either model of $w^{*}(s)$ and $v(s)$ is correct. 
\begin{theorem}[Double Robustness]\label{thm:dr_value}
Suppose $\forall k\leq K$, for some $w^{*\dagger},v^{\dagger}$, $\|\hat w^{*(k)}(s)-w^{*\dagger}(s)\|_2=\op(1),\|\hat v^{(k)}(s)-v^{\dagger}(s)\|_2=\op(1)$. Then, as long as $w^{*\dagger}=w^{*}$ or $v^{\dagger}=v$, $\hat J_{\mathrm{V2}}\stackrel{p}{\rightarrow}J$. 
\end{theorem}
In addition, the asymptotic MSE can be calculated under nonparametric rate conditions if the behavior policy is known. 

\begin{theorem}[Asymptotic Results]\label{thm:jv_mse}
Suppose $\forall k\leq K$, $\|\hat w^{*(k)}(s)-w^{*}(s)\|_2=\alpha_1,\|\hat v^{(k)}(s)-v(s)\|_2=\beta$, where $\alpha_1\beta=\op(n^{-1/2}),\alpha_1=\op(1),\beta=\op(1)$. Then, $\sqrt{n}( \hat J_{\mathrm{V2}}-{J})\stackrel{d}{\rightarrow} \bN(0,\Upsilon_{\TI2})$. 
\end{theorem}
Again note that this asymptotic MSE is larger than the efficiency bound in the pre-specified evaluation policy case \citep{KallusNathan2019EBtC}.

\end{document}